\documentclass[times,twocolumn,final,authoryear]{elsarticle}

\usepackage{ycviu}
\usepackage{framed,multirow}

\usepackage{amssymb}
\usepackage{latexsym}
\usepackage{amsmath, bm}

\usepackage{url}
\usepackage{xcolor}
\definecolor{newcolor}{rgb}{.8,.349,.1}

\journal{Computer Vision and Image Understanding}

\begin{document}

\clearpage

\ifpreprint
  \setcounter{page}{1}
\else
  \setcounter{page}{1}
\fi

\begin{frontmatter}

\title{GMC: A General Framework of Multi-stage Context Learning and Utilization for Visual Detection Tasks}

\author[1]{Xuan \snm{Wang}\corref{cor1}}
\ead{xwang4@gradcenter.cuny.edu}
\author[1,2]{Hao \snm{Tang}} 
\author[1,3]{Zhigang \snm{Zhu}} 
\cortext[cor1]{Corresponding author}

\address[1]{The Graduate Center, The City University of New York, 365 5th Avenue, New York, NY 10016, USA}
\address[2]{The Borough of Manhattan Community College, The City University of New York, 199 Chambers Street, New York, NY 10007, U.S.A}
\address[3]{The City College, The City University of New York, 160 Convent Avenue, New York, NY 10031, USA}

\received{1 May 2013}
\finalform{10 May 2013}
\accepted{13 May 2013}
\availableonline{15 May 2013}
\communicated{S. Sarkar}

\begin{abstract}
Various contextual information has been employed by many approaches for visual detection tasks. However, most of the existing approaches only focus on specific context for specific tasks. In this paper, GMC, a general framework is proposed for multistage context learning and utilization, with various deep network architectures for various visual detection tasks. The GMC framework encompasses three stages: preprocessing, training, and post-processing. In the preprocessing stage, the representation of local context is enhanced by utilizing commonly used labeling standards. During the training stage, semantic context information is fused with visual information, leveraging prior knowledge from the training dataset to capture semantic relationships. In the post-processing stage, general topological relations and semantic masks for stuff are incorporated to enable spatial context reasoning between objects. The proposed framework provides a comprehensive and adaptable solution for context learning and utilization in visual detection scenarios. \textcolor{black}{The framework offers flexibility with user-defined configurations and provide adaptability to diverse network architectures and visual detection tasks, offering an automated and streamlined solution that minimizes user effort and inference time in context learning and reasoning.} Experimental results on the visual detection tasks, for storefront object detection, pedestrian detection and COCO object detection, demonstrate that our framework outperforms previous state-of-the-art detectors and transformer architectures. 
The experiments also demonstrate that three contextual learning components can not only be applied individually and in combination, but can also be applied to various network architectures, and its flexibility and effectiveness in various detection scenarios. 
\end{abstract}


\end{frontmatter}


\section{Introduction}
\label{introduction}

Contextual information plays a significant role in various computer vision tasks, encompassing both visual and non-visual data related to the appearance of a target, be it an object or an event. When objects are encountered without proper context, such as in object recognition, the task can become challenging. However, leveraging contextual cues can offer vital insights for accurate target recognition. In tasks involving videos, like action or event recognition, temporal context becomes crucial in predicting future occurrences. For instance, if a person walking is partially obscured by a car or a telegraph pole in the current frame, information from adjacent frames (previous or next) can aid in locating and detecting the occluded person.

In object detection tasks, the presence of other objects within the scene can influence the identification of a target object. These contextual cues can reveal co-occurrences and object locations. For instance, a painting should typically be found on a wall rather than on the ground. Knowing that there is a desktop on a table increases the likelihood of finding a keyboard and a mouse nearby. Furthermore, additional contextual information such as locations, dates, and environments can further enhance the likelihood of detecting objects or events. 

\begin{figure}[ht]
\begin{center}
   \includegraphics[width=1\linewidth]{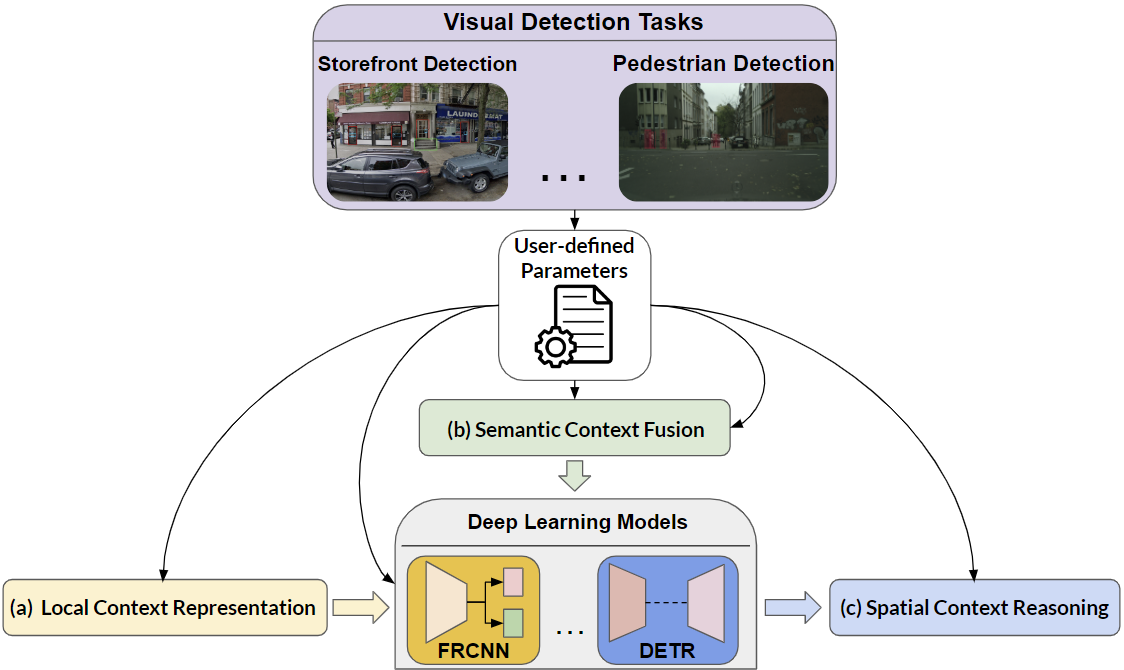}
\end{center}
   \caption{The overview of GMC, our general framework of multi-stage context learning and utilization for visual detection tasks. We design a user configuration mechanism for automating the process for various detection tasks and with different network models. Each context component is guided by user-defined parameters with minimum modification of the system when applying to different deep learning models and visual tasks.}
\label{fig:framworkoverview}
\end{figure}

A comprehensive survey on context understanding in computer vision can be found in our recent survey paper \citep{wang2023context}.
In this paper,we propose a \textit{G}eneral framework of \textit{M}ulti-stage \textit{C}ontext learning utilization (the \textit {GMC framework}) for visual detection tasks. The GMC framework incorporates different forms of contextual information, works for different visual detection tasks, and can use different network architectures (Fig. \ref{fig:framworkoverview}). The forms of context information include local context in the data labeling stage, semantic context in the model training stage, and spatial context among objects to be detected in the post-processing stage. This framework aims to offer the generality of using context in various tasks and with various architectures, in order to improve performance in various visual detection tasks.

In the domain of visual object detection, bounding boxes are widely used to represent the spatial location of objects. Crowdsourcing platforms like Amazon's Mechanical Turk (AMT) are commonly employed to annotate large datasets such as MSCOCO \citep{lin2014microsoft} and ImageNet \citep{deng2009imagenet}, heavily relying on human labelers. Typically, human labelers manually draw tight bounding boxes around objects to maintain label consistency. However, when dealing with small objects, using tight bounding boxes may not provide sufficient local contextual information for accurate recognition. In some cases, even human observers struggle to recognize small objects due to their small sizes. Moreover, viewing the entire scene allows for even easier recognition by incorporating a more global context, despite the size of a small object in the image.

Previous studies such as \citep{lim2021small, leng2021realize} have demonstrated the importance of contextual information from the surrounding areas of small objects in achieving successful detection results. However, these studies typically utilize deep learning models to extract and refine features from these small objects, which can increase computational costs. In fact, one straightforward approach to leverage local context for small objects is to directly include their surrounding areas in the images during the labeling process, thereby providing explicit contextual information.

In this work, we propose an automatic \textit {local context} representation that enhances the original bounding boxes for specific objects. This allows us to incorporate local context prior to the model training step, by simply using the two most commonly used definitions of small objects in computer vision tasks. By adopting this approach, we aim to exploit the benefits of local context while mitigating the potential increase in computational complexity associated with deep learning-based feature extraction methods.

\textit{Semantic context} plays a crucial role in successful object detection by providing valuable information. Even without visual cues, knowing that a scene is set in an urban street environment allows us to make educated guesses about the presence of pedestrians, bicycles, vehicles, and other relevant objects. The labels assigned to objects within a scene in a training dataset can also provide prior knowledge regarding the co-occurrence relationships between different labels. Previous studies such as \citep{li2014multi, li2016conditional, lee2018multi} have demonstrated the effectiveness of using graphs to model label correlations. For instance, Chen et al. \citep{chen2019multi} proposed a framework that leverages graph-based label dependencies for multi-label image recognition. Wang et al. \citep{wang2022multiclu} model the highly correlated storefront objects using the co-occurrence of the related objects and leverage the context information for better detection performance.

Inspired by these approaches, we extend the idea we proposed in \citep{wang2022multiclu} for storefront accessibility detection, and introduce a mechanism that allows easy user configuration to automate the generation of a contextual graph and the retrieval of word embeddings from pre-trained language models. This mechanism enables the adaptation of context learning models to various visual detection tasks. Within our framework, a Graph Convolutional Network (GCN) \citep{kipf2016semi} is utilized to learn from the contextual graph. By incorporating word embeddings, the GCN builds a semantic space and projects visual features extracted by the object detector into this space for the final classification stage. This integration of semantic context enhances the accuracy and performance of the object detection system.

Real-world scenes often exhibit spatial relationships between objects (i.e., \textit{spatial context}), where certain objects tend to appear together or have specific spatial arrangements. For instance, a keyboard and a mouse are commonly found together, with the mouse typically positioned to the right of the keyboard. Yang et al. \citep{yang2015facial} proposed a Faceness-Net that leverages spatial relationships between facial parts, such as the hair appearing above the eyes and the nose appearing below the eyes. Similarly, another work \citep{yang2019step} introduced a spatial-aware network that models relative locations among different objects to improve object detection performance. Recent papers \citep{wang2022multiclu, chacra2022topology} have also utilized specific spatial relationships for tasks like storefront accessibility detection and scene graph generation. However, these methods often employ hard-coded spatial relationships tailored to their specific tasks, making it challenging to generalize them to other tasks without significant modifications. 

To address this limitation and provide a more general approach to model spatial relationships, topological relationships can be beneficial for capturing object relations, as shown in Fig. \ref{fig:framworkoverview}. 
In this work, we extended the idea and propose a more generalized approach to model spatial relationships between objects for visual detection tasks. By utilizing a user configuration mechanism, we maximize flexibility in defining object relations without the need for code modifications.

While contextual information has been employed in specific computer vision tasks, such as data augmentation \citep{dvornik2018modeling}, semantic reasoning during training \citep{zhu2021semantic, chen2019multi, wang2022multiclu, visapp23wang}, and post-processing \citep{fang2017object, wang2022multiclu, visapp23wang}, there is a lack of research on a comprehensive general framework that guides context learning across data labeling, model training, and post-processing stages in a generalized manner. In our previous work \citep{wang2022multiclu}, we proposed a context learning framework for storefront accessibility detection that covered these stages. However, the framework was specifically designed with context learning mechanisms tailored to storefront accessibility detection. Therefore, significant code modifications were necessary to adapt it to different tasks. In a follow-up work \citep{visapp23wang}, we proposed a framework for different visual detection tasks in urban scenes. However the framework only works with one single network architecture, and the experiment on the second example (the pedestrian detection) is very limited; There are no available contexts for spatial context reasoning and the result only achieves minor improvement over the baseline network. 

In this work, we present a general context learning and reasoning framework with various deep learning models, applicable to various visual detection tasks, and therefore offering greater flexibility and adaptability without requiring extensive code changes. As an extended version of our previous work \citep{wang2022multiclu, visapp23wang}, this paper demonstrates the versatility and adaptability of our context components by successfully applying them to different deep learning models with minimal modification. The pedestrian detection task is greatly enhanced with more categories of contextual objects and includes all the three stages of context reasoning. \textcolor{black}{We also tested the framework on a large detection benchmark - MSCOCO dataset, showing promising results.}

The proposed context learning and reasoning framework for visual detection tasks offers several noteworthy aspects. Firstly, it introduces a comprehensive approach consisting of three key components: Local Contextual Representation (LCR), Semantic Context Fusion (SCF), and general Spatial Context Reasoning (SCR). The LCR component improves recognition accuracy for specific objects especially small objects by incorporating their local context, while the SCF component models semantic relations using a contextual graph, capturing co-occurrence and contextual dependencies. Additionally, the SCR component leverages topological relationships and semantic masks to incorporate general spatial relations between objects. The framework's flexibility allows for easy adaptation to different tasks, without requiring extensive code modifications. Overall, this framework presents a valuable contribution to the field of visual detection by providing a comprehensive and adaptable solution that enhances context learning and reasoning capabilities.

 \textcolor{black}{As some highlights,the local context representation and semantic context fusion components are seamlessly integrated into diverse models, ensuring an automated adaptation process in using ground truth labels and prior knowledge. This integration is also designed to empower users with the flexibility to tailor the components according to their specific requirements through the utilization of user-defined parameters. Moreover, we have introduced a novel general spatial context reasoning component that combines topological relations between objects and semantic masks. This combination allows our framework to easily adapt to various visual detection tasks, providing a powerful tool for improving detection performance in diverse scenarios with more accurate results. We provide user flexibility to configure the spatial relations because the user configuration can offer meaningful definitions of important spatial relations as the first step, and then our Spatial Context Reasoning (SCR) component will autonomously generate relation parameters, such as overlapping thresholds based on the provided information in the user configuration, by modeling subject-object ground truth labels.} Overall, our approach not only enhances the effectiveness of context learning and reasoning in visual detection but also simplifies the integration process, making it readily applicable to a wide range of deep learning models and tasks.

In summary, the main contributions of this paper are:
\begin{itemize}
    \item We introduce a general framework for multistage context learning and utilization, with three context components to leverage local context, semantic context and spatial context. This combination of components provides a holistic solution to address context learning and reasoning in visual detection tasks.
    \item Our framework proposed in this work is designed to be applicable to any deep learning models. This versatility makes the framework highly versatile and empowers users to leverage its benefits across different object detection tasks, regardless of the specific deep learning model employed.
    \item Our framework is not limited to a specific visual detection task but can be applied to various visual detection tasks, including storefront object detection and pedestrian detection as our examples. Its flexibility and adaptability enable users to utilize the framework across a wide range of visual detection tasks, benefiting from its context learning and reasoning capabilities.
    \item Our framework has the ability to incorporate different types of context information at various stages of the detection process. It provides a unified framework that can effectively integrate and utilize these contextual cues at the appropriate stages, such as during data preprocessing, model training, or post-processing. This capability enhances the overall performance and robustness of the detection system by harnessing diverse sources of contextual information to improve object understanding and localization.
\end{itemize}

The paper is organized as follows. Section \ref{relatedwork} discusses related work. Section \ref{componentdesign} proposes our general context learning and reasoning framework and describes each component in detail. Section \ref{networkarchitectures} discuss how the general framework work with various deep learning network architectures with minimal modification of the code. Section \ref{tasksandexperiments} discuss the use of the general framework for three different tasks, including a description of the three datasets (Section \ref{datadescription}), and the experimental results (Section \ref{experimentresults}). Finally, Section \ref{conclusion} provides a few concluding remarks.

\section{Background and Related Work} \label{relatedwork}

In this section, we will start with a general survey of the literature in context learning and utilization for computer vision tasks, then move on the use of context information in object detection, and finally focus on pedestrian detection - a particularly important task that poses challenges and opportunities in using context information.

\subsection{Context Learning and Utilization}

Humans use visual context effortlessly to perceive the real world. An object hanging on the wall is probably a painting, not a car. A doorknob should be within the frame of a door, not on the ground. Contextual information provides critical information to help us visually find and recognize objects faster and more accurately. Not only in human perception, contextual information also plays an important role in many computer vision tasks, such as object detection \citep{du2012context, fang2017object, sun2017seeing, zhu2016could, zhu2021semantic}, video event recognition \citep{wang2015video, wang2016hierarchical}, video action detection \citep{yang2019step, zhu2013context}, scene graph generation \citep{xu2017scene, zellers2018neural}, data augmentation \citep{dvornik2018modeling}, image classification \citep{mac2019presence}, and image inpainting \citep{pathak2016context}. In these tasks, different forms of  contextual information have been employed. The contextual information used in the literature includes: global context \citep{zellers2018neural}, local neighborhood context \citep{pathak2016context, dvornik2018modeling, du2012context}, prior semantic knowledge \citep{wang2015video, wang2016hierarchical}, geographic information \citep{mac2019presence}, spatial relation between objects \citep{sun2017seeing, xu2017scene, zellers2018neural, yang2019step} and temporal information \citep{wang2015video, wang2016hierarchical, yang2019step, zhu2013context}.

Context information has been widely used in many computer vision tasks. Dvornik et al. \citep{dvornik2018modeling} show that the visual context surrounding objects is crucial to predict the presence of objects. A serial work \citep{wang2015video, wang2016hierarchical} introduces a hierarchical context model to recognize events in videos. Wang et al.\citep{wang2022multiclu} make use of various contextual information by applying a unified multi-stage framework in context learning and utilization from data labeling, model training, to object detection and result evaluation.

Context has been integrated in different ways in visual detection tasks. Many visual detection tasks \citep{yang2015facial, chen2019multi, pathak2016context, leng2021realize, li2016human, mac2019presence, yang2018graph} implement context information into the backbone models and aggregate with the features extracted from context-free methods. Deep learning methods mainly have four stages: data pre-processing (including labeling), model training, post-processing, and result evaluation. Context information has either been aggregated during the training stage or used in the post-processing stage. No general pipelines have been proposed on how we can incorporate context through the whole process stages. Although different context integration can be used in a single stage or in multiple stages, a general pipeline is needed to guide the integration for context. Our proposed framework employs different forms of context information through the entire deep learning process, and each component is easy to add and remove from an object detector.

\subsection{Object Detection}
Contextual information plays a crucial role in understanding natural scenes and images for object detection, as it provides rich information about the relationships between objects and the overall scene. However, the evaluation of context models has primarily focused on improving object detection performance for particular tasks, overlooking more general applications of contextual information. In the domain of urban scene object detection, various methods have been proposed, addressing specific tasks such as text detection and recognition \citep{du2012context, zhu2016could}, zebra crossing detection \citep{ahmetovic2015zebra}, curb detection \citep{cheng2018curb, sun2017seeing}, and storefront accessibility detection \citep{wang2022multiclu}.

For example, Du et al. \citep{du2012context} and Zhu et al. \citep{zhu2016could} focused on text detection in street environments. Cheng et al. \citep{cheng2018curb} proposed a framework for road and sidewalk detection using stereo vision in urban regions. Sun et al. \citep{sun2017seeing} aimed to identify missing curb ramps at street intersections by leveraging the pairwise existence of curb ramps. Our recent work \citep{wang2022multiclu} introduced a multi-stage context learning framework specifically designed for storefront accessibility detection, utilizing category-specific relations. These examples demonstrate that context modeling has been applied to various urban scene object detection tasks beyond traditional object recognition. It highlights the potential of exploiting different types of contextual information to improve the performance of detection systems in diverse real-world scenarios. In this paper we propose a general context learning and reasoning framework which could be adapted to various visual detection tasks.

Contextual information, particularly prior knowledge, has played a crucial role in advancing object detection tasks. Fang et al. \citep{fang2017object} introduced a knowledge-aware object detection framework that incorporates external knowledge, such as knowledge graphs, into object detection algorithms. By leveraging a knowledge graph, which represents real-world concepts and their interactions, this framework enables the modeling of semantic consistency. Even concept pairs that are not directly connected in the graph can benefit from this approach, leading to enhanced generalization capabilities.

Similarly, Zhu et al. \citep{zhu2021semantic} explored the integration of semantic context and visual information for the task of few-shot object detection. Their work focused on explicit relation reasoning and utilized word embeddings to represent class labels. By establishing semantic relation consistency between base and novel classes, the aim was to bridge the domain gap between visual and language information. Incorporating semantic consistency principles, their framework improved object detection by optimizing for better alignment with prior knowledge.

Building upon these concepts, our general framework embraces the notion of semantic consistency to quantify and generalize knowledge, resulting in improved object detection performance through a re-optimization process. In addition, our framework adopts a context-aware approach to object detection, considering both visual context and prior knowledge context. By incorporating both types of context, our framework provides a more comprehensive and enriched understanding of the scene, leading to more accurate and robust object detection results.

Indeed, context can be leveraged not only for detecting objects but also for predicting their presence or absence in an image. Sun \citep{sun2017seeing} conducted a unique vision task focused on identifying the absence of objects in an image, specifically curb ramps. This work extensively utilized local and spatial context information to determine the locations where curb ramps should exist.

Similarly, in our proposed framework, we emphasize the importance of local context representation surrounding small objects. This local context provides valuable information that can indicate both the location and category of the object. By incorporating this local context into our general framework, we aim to enhance the detection and prediction capabilities, enabling more accurate understanding of the scene and object presence even in the absence of explicit object instances.

\subsection{Pedestrian Detection}

Pedestrian detection in urban scenes presents unique challenges due to factors such as heavy occlusion and small-scale pedestrian images. Several papers have focused on addressing these challenges and improving the performance of pedestrian detection algorithms. For example, Cai et al. \citep{cai2016unified} proposed a unified framework for pedestrian detection that incorporates contextual information to handle occlusion. Zhang et al. \citep{zhang2017citypersons} introduced the CityPersons dataset specifically for pedestrian detection in urban environments and proposed a scale-aware network to tackle the problem of detecting small-scale pedestrians.

Other works have explored different approaches to handle occlusion in pedestrian detection. Zhou et al. \citep{Zhou_2018_ECCV} proposed an attention-based method that focuses on visible parts of partially occluded pedestrians, improving the detection accuracy in challenging scenarios. Wu et al. \citep{wu2020mimic} introduced a part-based detection framework that leverages feature transformation to handle occlusion and improve detection performance.

Despite the progress made by CNN-based pedestrian detectors, there are still limitations in detecting small-scale and heavily occluded pedestrians. These challenges require further exploration and innovation in the design of detection algorithms. For example, the integration of additional context information beyond a single image, such as global scene context and temporal context, could potentially improve the performance of pedestrian detection systems in real-world scenarios. This is beyond the scope of this paper; more details can be found in our recent survey paper\citep{wang2023context}.

Pedestrian detection in urban scenes is a challenging task that has garnered significant attention in the computer vision community. Several papers have focused on addressing the unique challenges associated with detecting pedestrians in such environments. While approaches like Faster R-CNN have become popular for pedestrian detection, they often fall short in effectively handling heavily occluded pedestrians and small-scale pedestrians. Limited progress has been made in leveraging local context information specifically for these scenarios, resulting in sub-optimal detection performance.

To address this gap, our proposed novel framework integrates local context for small-scale and occluded pedestrian detection in urban scenes. Our approach incorporates general topological relations among objects to facilitate spatial reasoning. By considering the relationships (including occlusions) between objects, we can reason about the presence and location of pedestrians, even in challenging situations. Notably, our framework goes beyond improving pedestrian detection alone; it also enhances the detection results for other objects in the scene. By leveraging the synergistic effects of contextual components, our approach aims to achieve superior performance compared to existing methods. 

By emphasizing the importance of local context and introducing general topological reasoning, our framework offers a comprehensive solution for pedestrian detection in urban scenes. Note that the general framework is not specially designed for pedestrian detection but the system can be configured to tackle these two challenges in pedestrian detection.
Through the incorporation of contextual cues and the utilization of interplay between different components, we can overcome the limitations of traditional approaches and improve detection accuracy. Ultimately, our work contributes to advancing the understanding of urban scenes and objects, opening up new possibilities for real-world applications.

\begin{figure*}[ht]
\begin{center}
   \includegraphics[width=1\linewidth]{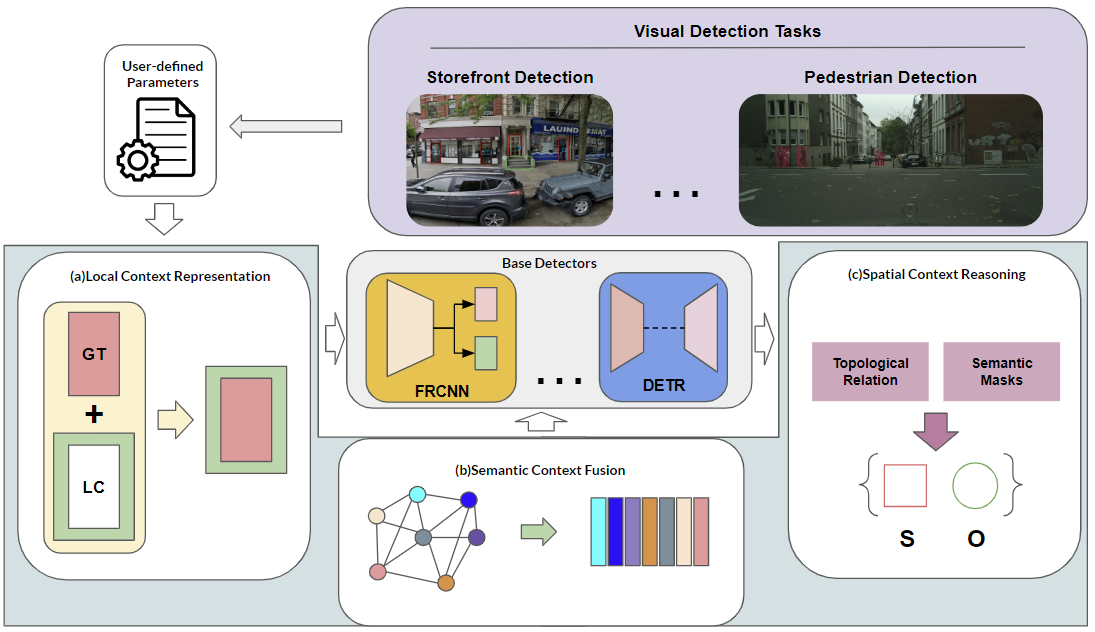}
\end{center}
   \caption{Details of our GMC framework, the general framework of multi-stage context learning and utilization for visual detection tasks. We design a user configuration mechanism for automating the process for various detection tasks (e.g., storefront object detection, pedestrian detection), using different base detectors (e.g. a CNN model Faster R-CNN (FRCNN) and a transformer model DETR. Three context learning and utilization components - (a) Local Context Representation, (b) Semantic Context Fusion, and (c) Spatial Context Reasoning, guide the deep learning models during data labeling, model training and post-processing stages. Each component can be applied individually and in combination. \textit{GT}: Ground Truth. \textit{LC}: Local Context. \textit{S}: Subject. \textit{O}: Object.}
\label{fig:framworkdetail}
\end{figure*}

\section{General Framework and Context Components} \label{componentdesign}

Our proposed GMC framework, as detailed in Fig. \ref{fig:framworkdetail}, consists of three key context components: local context representation, semantic context fusion, and spatial context reasoning. These components can be applied individually or in combination with a given visual detection network architecture to enhance object detection performance. 

The local context representation component (Section \ref{lcr}) focuses on capturing local contextual information specific to the objects of interest. By incorporating local context features in the data labeling stage, this component improves the accurate detection of objects, particularly small-scale or occluded ones, by leveraging relevant contextual cues. The semantic context fusion component (Section \ref{scf}) integrates semantic information with visual context to capture object relationships. By combining prior knowledge and/or learning from the training dataset in the model training stage, this component enhances the detection network's understanding of the scene and improves its ability to discriminate and classify objects. The spatial context reasoning component (Section \ref{scr}) introduces a general topological relation between object categories to optimize detection results. By considering the spatial relationships between objects in the post-processing stage, such as "above," "under," or "within," this component refines detection outputs based on their spatial arrangements. This spatial context reasoning enhances the detection network's localization accuracy and object classification performance by incorporating topological reasoning into the detection process.

\textcolor{black}{An automated process is implemented for each component with simple user defined parameters. In local context representation component, we apply an automatic local contextual labeling approach to enhance the original bounding boxes for small objects in order to employ local context \textit{before} the model training step, by using the two most used definitions of small object in computer vision tasks. In semantic context fusion component, we automate the process for generating a contextual graph by leveraging label occurrence knowledge from training data, and automatically searching the word embeddings from a pretrained language model. In spatial context reasoning component, we adopt user configuration for important spatial relations of objects as guidelines, to automatically generate the spatial relation thresholds, which  maximize the flexibility for object relation definition, without code modifications.}

In the following sections, we will provide detailed explanations of each component within our proposed general framework. Through some user-defined parameters related to a given visual detection task and the chosen base detector, the GMC framework can be easily configured to form an end-to-end model for the task.

\subsection{Local Context Representation}
\label{lcr}

\begin{figure}[ht]
\begin{center}
   \includegraphics[width=1\linewidth]{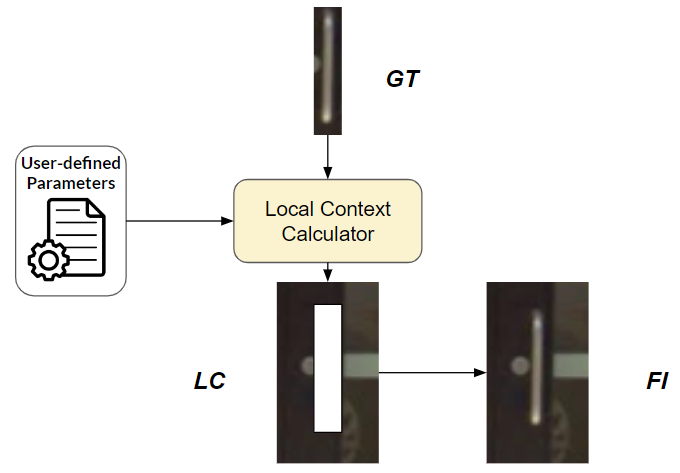}
\end{center}
   \caption{An utilized local context representation. The local context calculator is guided by user-defined parameters and enhance the local context around the ground truth label of the object. \textit{GT}: Ground Truth. \textit{LC}: Local Context. \textit{FI}: Final Input.}
\label{fig:lcrexample}
\end{figure}


The concept of \textit {local context} for objects, particularly small ones, takes center stage in the Local Contextual Representation (LCR) component. In the realm of computer vision, categorizing an object as "small" isn't always clear-cut. Factors like shooting angles and environmental conditions can render an object that's deemed "small," such as a spoon, appearing quite "large" within an image. Hence, the notion of smallness hinges on an object's size relative to the context of the image, as explained further below. The procedural essence is graphically illustrated in Figure \ref{fig:lcrexample}. A local context calculator is at the heart of this process, guided by user-defined parameters specific to LCR. This calculator works to enrich the local context surrounding the ground truth label of the targeted object. To initialize this local context calculator, we introduce two commonly embraced standards for characterizing small objects. \textcolor{black}{The Local Context Representation (LCR) component operates during the data preprocessing stage, focusing solely on the labeling standard and the specified enlargement percentage for small objects (Table \ref{tab:parametersummary}). This component automatically processes the labels before they are fed into the network, ensuring seamless integration without introducing additional inference complexity.}

Within the COCO dataset \citep{lin2014microsoft}, small objects are defined as those whose dimensions are $32 \times 32$ pixels or smaller, within the confines of an image with a fixed size of $640 \times 480$ pixels. Another definition, as detailed in \citep{chen2016smallobject}, relates to situations where the overlap area between the ground truth bounding box and the image remains below 0.58\%. Given the robustness and widespread adoption of these definitions in the research community, we employ them as reference points for automating the labeling process for small objects. We include the surrounding local context of the bounding box $B$ of an object $O$ in image $I$ if the object satisfies with the COCO standard for a small object as:
\begin{equation}
\label{eq:cocostandard}
    B'_{O}=
    \begin{cases}
      (1+\alpha) B_{O}, & \text{if}\ B_{O} < 32 \times 32  \\
      B_{O}, & \text{otherwise}
    \end{cases}
\end{equation}
If the small object satisfies with the second standard - the Small Object Dataset (SOD) Standard \citep{chen2016smallobject}, we include the local context of the bounding box $B$ of the object $O$ in image $I$ by:
\begin{equation}
\label{eq:sodstandard}
    B'_{O}=
    \begin{cases}
      (1+\beta) B_{O}, & \text{if}\ \frac{B_{O}}{R_{I}} < 0.58\%  \\
      B_{O}, & \text{otherwise}
    \end{cases}
\end{equation}
The above equations introduce notations representing the original and updated bounding boxes of the ground truth label for a small object. These notations, $B_{O}$ and $B'_{O}$ respectively, are utilized in the context of the user-defined parameters for the Local Context Representation (LCR) component. Firstly, the parameters $\alpha$ and $\beta$ hold significance as extending factors, expressed in terms of a percentage, from the original bounding boxes. These factors are related to two distinct standards: the COCO standard and the SOD standard. The resolution of the input image, denoted as $R_{I}$, is automatically determined. This automatically calculated resolution serves as a crucial component in the calculation of these factors. Secondly, the framework affords users the liberty to choose between the two contextual labeling standards. Should a given small object meet the criteria of both definitions, the user can opt for the standard that best aligns with their requirements. \textcolor{black}{Importantly, both the original bounding boxes and the enlarged bounding boxes are retained for all small objects that conform to the user-selected standard for both training and testing sets.} This dual retention strategy serves the dual purpose of integrating local contextual information and enhancing the detection's robustness. The forthcoming sections will delve into the specifics of the experimental settings in Section \ref{experimentsettings}, elaborating further on these parameters and their implications.


\subsection{Semantic Context Fusion}
\label{scf}

Semantic information indeed plays a crucial role in visual detection tasks, providing valuable insights to enhance the detection process. To ensure a seamless and automatic Semantic Context Fusion (SCF) into our framework, we have introduced the SCF user-defined parameters, namely, the categories of a given visual detection task and the text embeddings used in the task. For example, for a storefront object detection task, they are door, doorknob, stair. For pedestrian detection, they include pedestrian, vehicle, bicycle (bike), motorcycle, etc. These parameters act as guiding factors for the model to learn and incorporate semantic context using text embeddings. The text embeddings, obtained from pre-trained language models, are utilized to generate semantic spaces that can be effectively fused with the visual information obtained from the detection process. This integration of semantic context with text embeddings allows our framework to automatically leverage valuable semantic information to improve the overall detection performance, while minimizing the need for extensive component modification.

\begin{figure}[ht]
\begin{center}
   \includegraphics[width=1\linewidth]{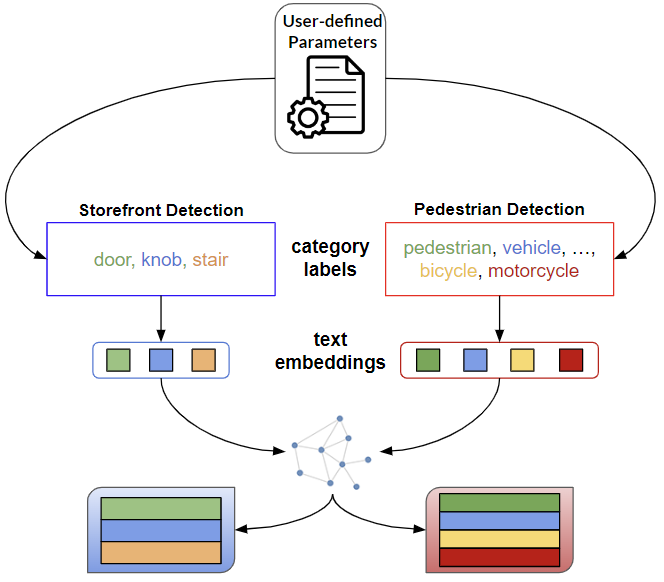}
\end{center}
   \caption{The visualization of Semantic Context Fusion. We use category information as the semantic context cues to generate semantic spaces for visual detection tasks.}
\label{fig:SCFcomponent}
\end{figure}

In our framework, the fusion of semantic context is depicted in Figure \ref{fig:SCFcomponent}. When the framework receives category information from the SCF user configuration, it proceeds to search for word embeddings $H_{labels} \in \mathbb{R}^{n \times d}$ from a pretrained language model (such as GloVe \citep{pennington2014glove}). Here, $n$ represents the number of label categories, and $d$ denotes the dimensionality of the word embeddings. Subsequently, an automatic generation of the contextual graph takes place. The Graph Convolutional Network (GCN) is then employed to learn semantic relations within the contextual graph, effectively constructing a semantic space. This semantic space is obtained by transforming the label feature representation, resulting in ${H'}_{labels} \in \mathbb{R}^{n \times D}$, where $D$ represents the dimensionality of the region features extracted from the object detector. As illustrated in Figure \ref{fig:framworkdetail}, the region features $f_{regions} \in \mathbb{R}^{D \times N}$ are projected into the semantic spaces ${H'}_{labels}$. Ultimately, the final output is derived from this process:

\begin{equation}
            \textbf{P}_{regions} = softmax({H'}_{labels}{f}_{regions})
\end{equation}
where $\textbf{P}_{regions}$ represents the classification probability distribution for each proposed region, and $\textbf{P}_{regions} \in \mathbb{R}^{n \times N}$.

\textcolor{black}{As the category information is provided by a given task, our system automatically generates a contextual graph between different categories, leveraging prior label occurrence knowledge extracted automatically from the training data. Additionally, we autonomously search for pretrained word embeddings from the dictionary \citep{pennington2014glove} without requiring extra information. The SCF (Semantic Context Fusion) component, armed with the prebuilt contextual graph and pretrained word embeddings, ensures minimal additional complexity. The user-defined parameters for the SCF module are detailed in Table \ref{tab:parametersummary}.}

\subsection{Spatial Context Reasoning}
\label{scr}

\begin{figure}[ht]
\begin{center}
   \includegraphics[width=0.8\linewidth]{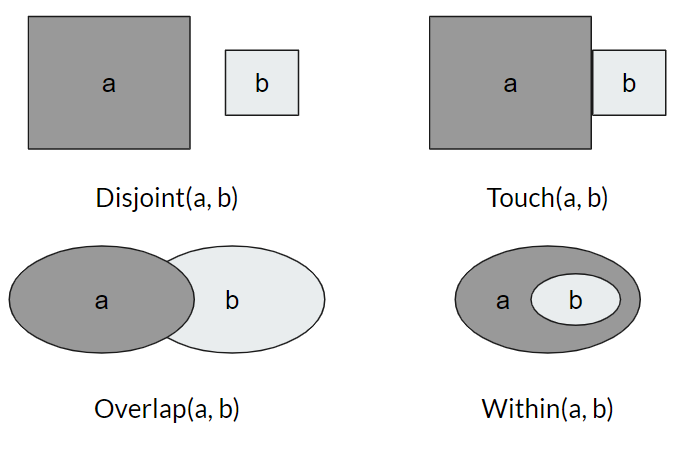}
\end{center}
   \caption{The visualization of common used topological relationships from \cite{clementini1993small} and \cite{egenhofer1991point}.}
\label{fig:topologicalvis}
\end{figure}

\begin{table*}[ht]
    \caption{Summary of the provided user-defined parameters for the  contextual components.}
    \begin{center}
    \resizebox{0.85\linewidth}{!}{
    \begin{tabular}{|l|c|c|}
    \hline
    Parameters & Context component & Definition \\
    \hline 
    $[\bm{Subject, Object}]$ & LCR\textbackslash SCR & Subject and object pair  \\
    \hline
    $\bm{Labeling\_standard}$ & LCR & The standard for small object label enlargement \\
    $\bm{Enlarge\_percentage}$ & LCR & The enlarging percentage for small object labels \\
    \hline
    \textcolor{black}{$\bm{Categories}$} & \textcolor{black}{SCF} & \textcolor{black}{The object categories} \\
    $\bm{Relation\_descriptor}$ & SCF & The contextual graph generation method \\
    \hline
    $\bm{pred}$(optional) & SCR & Directional relationships between subject and object \\
    $\bm{t}$ & SCR & Topological relationships between subject and object \\
    $\bm{Overlap\_threshold}$(optional) & SCR & The threshold of overlap percentage between subject and object \\
    $\bm{Search_{\bm{height}}}$(optional) & SCR & The height of search area for object \\
    $\bm{Search_{\bm{width}}}$(optional) & SCR & The width of search area for object \\
    \hline
    \end{tabular}
    }
    \end{center}
    \label{tab:parametersummary}
\end{table*}

In the proposed general Spatial Context Reasoning (SCR) component, we leverage topological relationships to model the spatial relations between different objects. Topological relationships provide a general and abstract representation of the relationships between objects, such as \textit{overlap}, \textit{within}, \textit{touch}, and so on. These relationships capture the overall spatial configuration and arrangement of objects in a scene, including next two each other, within, and occlusion. The visualization of topological relationships is depicted in Fig \ref{fig:topologicalvis}, illustrating how different objects can be related in terms of their spatial positions and co-occurrence. By incorporating topological reasoning, our framework enables a more comprehensive understanding of the spatial context, enhancing the object detection performance and facilitating richer semantic interpretations of the scene.

\begin{figure}[ht]
\begin{center}
   \includegraphics[width=1\linewidth]{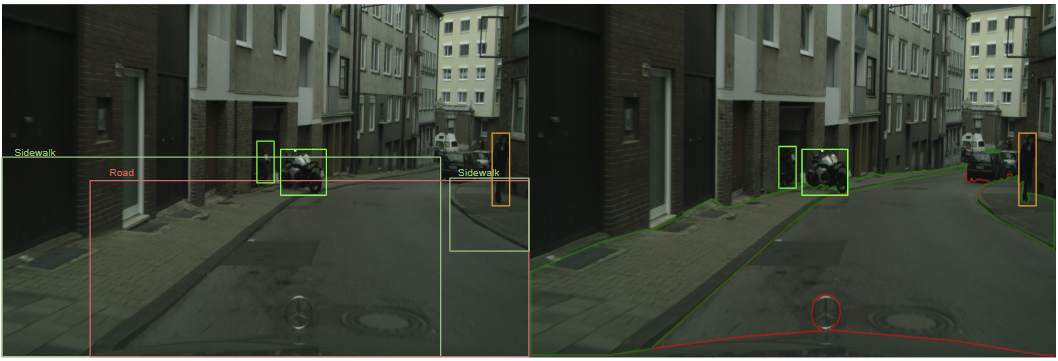}
\end{center}
   \caption{Bounding box vs. semantic masks for road and sidewalk.}
\label{fig:bbvsmask}
\end{figure}

We utilize a predicate $pred$, such as \textit{above}, \textit{under}, etc., to describe the directional relation between a subject and object pair [$S$, $O$], along with the topological relationship $t$, such as \textit{overlap} and \textit{within}. This general relation $R$ is defined as shown in Equation \ref{eq:topoeq}: 
\begin{equation}
\label{eq:topoeq}
            R[S, O] = pred[t(S, O)]
\end{equation} 
For instance, in urban settings, a common spatial relationship is that a stair is usually located under a door, even if there might be overlaps or spatial misalignment between them. The general relationship between a pedestrian and sidewalk can be described as $R[pedestrian, sidewalk] = under[overlap(pedestrian, sidewalk)]$. It is important to note that the general spatial relation is inversible, meaning that a pedestrian is on the sidewalk, and sidewalk can be considered under a pedestrian. To effectively apply this spatial reasoning, we define a search area around the detected subject, and if an object is detected within this search area and satisfies the condition defined by Equation \ref{eq:topoeq}. We propose it as a detection and send it for evaluation. In cases where multiple objects are detected within the search area, we propose the object with the highest score as the final prediction.

\begin{figure}[ht]
\begin{center}
   \includegraphics[width=1\linewidth]{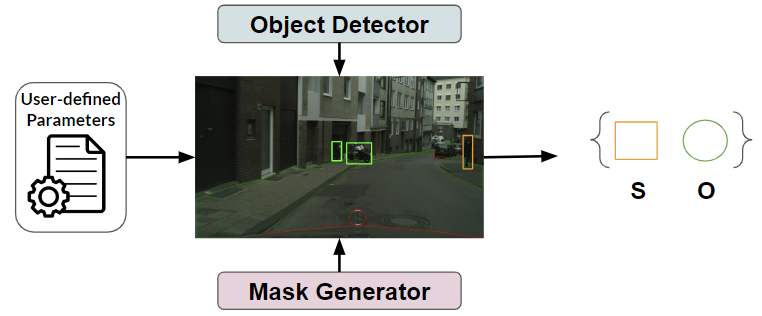}
\end{center}
   \caption{The visualization of general Spatial Context Reasoning.}
\label{fig:screxample}
\end{figure}

\begin{figure*}[ht]
\begin{center}
   \includegraphics[width=0.95\linewidth]{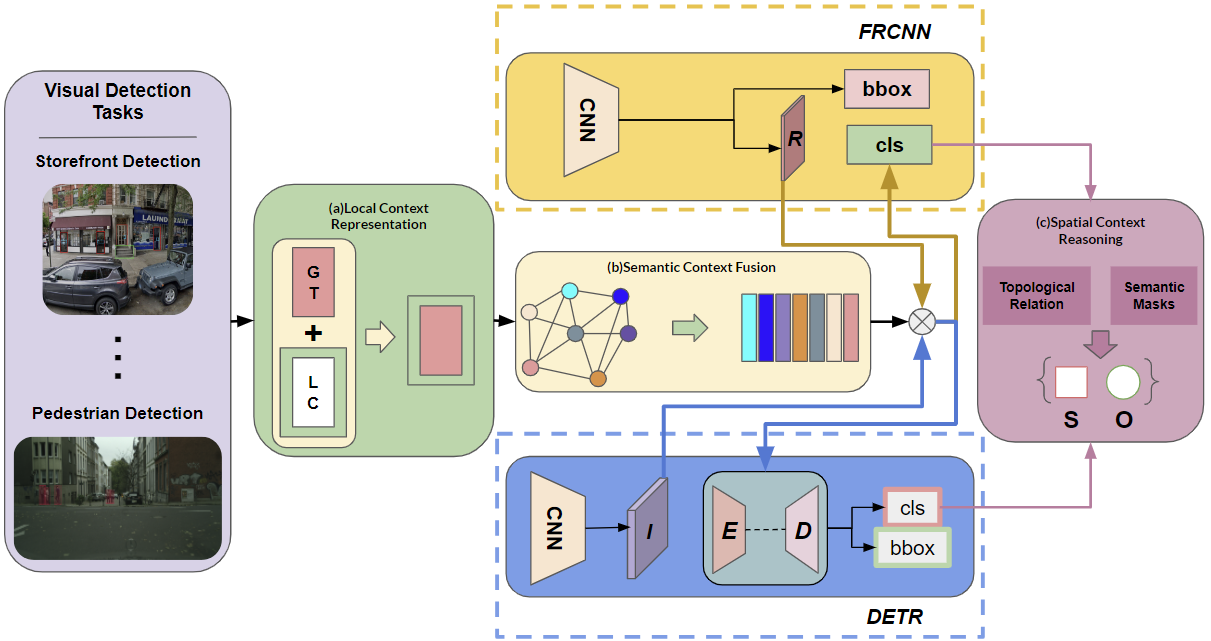}
\end{center}
   \caption{Integration of contextual components with different deep learning network architectures: Faster R-CNN (FRCNN) and DETR. \textit{GT}: Ground Truth; \textit{LC}: Local Context; \textit{S}: Subject; \textit{O}: Object; \textit{R}: Region features; \textit{I}: Image features; \textit{E}: Encoder; \textit{D}: Decoder; \textit{bbox}: bounding boxes; \textit{cls}: classification.}
\label{fig:variousnetworkoverview}
\end{figure*}

To enhance the applicability of our general framework to diverse visual detection tasks with more accurate detection, we have introduced \textit{semantic masks} in our general spatial context reasoning component. \textcolor{black}{As illustrated in Figure \ref{fig:bbvsmask}, bounding boxes for entities like roads and sidewalks may not be suitable for effective spatial reasoning between objects. In contrast, semantic masks offer a more precise and appropriate means for modeling the relationships between subjects and objects. While segmentation poses its challenges, modern state-of-the-art segmentation models can yield accurate masks for larger entities such as roads and sidewalks, rendering them readily usable for spatial reasoning.}  This addition allows us to segment large stuff such as sidewalks and roads using a pretrained model, which could significantly improves spatial reasoning in larger scenes. To measure the overlap between subject-object pairs, we use the intersection over subject (IoS) metric to describe the general spatial relation, as defined as: 

\begin{equation}
\label{eq:ioseq}
    IoS = \frac{(A_{s}\cap A_{o})}{(A_{s})}
\end{equation} 
where $A_{s}$ and $A_{o}$ denote the area of the subject and area of the object. The area can be bounding box or semantic mask based on the specific scenarios. This formulation enables us to capture the relative spatial arrangement of objects in a scene, which is valuable for improving the accuracy of object detection and localization across various visual detection tasks. We also provide users with the flexibility to configure the general spatial relation for the categories in their own dataset, allowing them to adapt the framework according to their specific task requirements. \textcolor{black}{
Moreover, the user configuration can offer meaningful definitions of important spatial relations as guidelines, and then our Spatial Context Reasoning (SCR) component will autonomously generate relation parameters such as overlap thresholds based on the information obtained from the ground truth labels. Through this adaptation, users can furnish general spatial relations for specific subject-object pairs. For instance, according to common sense, a car should be on the road, or a keyboard typically appears under the monitor. Using the provided relations, we automatically analyze the training dataset and establish overlap thresholds accordingly. This approach enables the model to leverage contextual information based on predefined spatial relationships, enhancing its understanding of the scene.} The user-defined parameters for LCR, SCF and SCR components are summarized in Table \ref{tab:parametersummary}.

\section{Working with Various Network Architectures}
\label{networkarchitectures}


The GMC framework can work with various deep learning network architectures with minimal modification of the code. In this paper, we give two examples, both which will be used in the tasks of our experiments. We employ two popular object detection frameworks, Faster R-CNN \citep{ren2015faster} and DETR \citep{carion2020end}, as the underlying detectors for both storefront accessibility detection and pedestrian detection tasks. These frameworks have demonstrated strong performance in various object detection scenarios. The integration pipeline of the three context components with Faster R-CNN and DETR is shown in Fig.~\ref{fig:variousnetworkoverview}. We will detail how the three context components can be seamlessly integrated with different backbone models, with minimal code modification.

Prior to the input of the visual detection task dataset into the model, we incorporate the Local Context Representation (LCR) component to augment the local context of specific objects. While we begin with two widely adopted definitions of small objects, as detailed in section \ref{lcr}, we also empower users to tailor the enhancement of local context according to their preferences by adjusting the enlarge percentage. This integration ensures that the LCR component can seamlessly adapt to diverse models without requiring any modifications to the underlying backbone models. This design approach not only increases the generality of our framework but also facilitates its ease of use and customization across different applications.

Within our Semantic Context Fusion (SCF) component, we harmonize semantic knowledge with visual features prior to the detection process. This integration is illustrated in Fig.~\ref{fig:variousnetworkoverview}. In the case of Faster R-CNN, we achieve this by mapping the extracted region features (R) from the feature extractor backbone into the semantic space, before subsequently feeding the resulting output into the classification (cls) head. In contrast, for a comparative scenario of DETR in Fig.~\ref{fig:variousnetworkoverview}, we first project image features (I) into the semantic space and subsequently input the resulting output into a transformer encoder-decoder (E\&D) for generating predictions. This design allows users to exercise control over the nature of the pretrained word embeddings in the SCF component, with the default setting being GloVe \citep{pennington2014glove}. The SCF component can be seamlessly integrated into each backbone architecture with minimal adjustments, signifying its adaptability and ease of incorporation into diverse models. This enables the enriched representation of contextual information in conjunction with visual cues, thereby enhancing the overall detection accuracy.

Moreover, the Spatial Context Reasoning (SCR) component can be seamlessly integrated to fine-tune the detected candidates by synergizing topological relationships and semantic masks among identified objects. The SCR component provides a valuable post-processing feature for both Faster R-CNN and DETR models, requiring minimal architectural adjustments. This adaptable SCR component can be easily integrated into the final stage of object classification (cls), offering a streamlined way to enhance object detection performance. Users retain the prerogative to exercise control over the component's parameters within the configuration file, ensuring adaptability and customization to distinct detection scenarios. This feature bolsters the accuracy of detection outcomes by leveraging not only the object-specific information but also the relationships and arrangements among objects within the scene.

\section{Tasks and Experiments}
\label{tasksandexperiments}


The general framework for context learning and utilization is designed not only for working with various visual detectors, but also for different tasks. In the following, we will showcase three examples: storefront accessibility detection, pedestrian detection, and COCO object detection. We will first introduce the three datasets, describe the experimental settings, and then detail the experimental results with the GMC framework.

\subsection{Dataset Description}
\label{datadescription}

\textbf{Storefront Accessibility Image Dataset.} For our experiments, we utilize the storefront accessibility image (SAI) dataset introduced in \citep{wang2022multiclu}. This dataset focuses on storefront accessibility in an urban environment and comprises three main categories: doors, knobs, and stairs. The SAI dataset is collected from Google Street View of New York City using the Google Street View API \citep{gsvapi}. To create the dataset, we employ the methodology described in \citep{Cavallo20153DCR} to compose panorama images. Each panorama image captures building facades on both sides of a street in New York City. Subsequently, we divide each formed panorama image into two halves, with each half covering one side of the facade. To ensure clear and easily labelable storefronts, we crop the center of each image, where contains the necessary visual information for storefront accessibility labeling.

\begin{figure}[ht]
  \centering
  \includegraphics[width=0.75\linewidth]{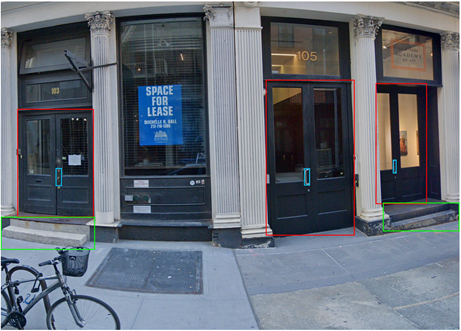}
  \caption{An example of labeled objects. Red: Ground truth bounding box of Door. Cyan: Ground truth bounding box of Knob. Green: Ground truth bounding box of Stair.}
  \label{fig:labelsample}
\end{figure}

\begin{table}[ht]
  \centering
  \caption{Statistics of collected storefront accessibility data.}
  \label{tab:datastatistics}
  \resizebox{0.75\linewidth}{!}{\begin{tabular}{|c|c|c|c|c|}
      \hline
        Dataset & \# of Images & Doors & Knobs & Stairs \\     \hline
        Train & 992 & 1885 & 1614 & 420 \\     \hline
        Test & 110 & 233 & 126 & 141 \\    \hline
  \end{tabular}
  }
\end{table}

The SAI dataset consists of a total of 1,102 images, where each image has been labeled for three main categories of accessibility: Door, Knob, and Stair. The labeling process was carried out using the Labelbox platform \citep{sharma2019labelbox}. To split the dataset for training and testing, a random sampling technique was employed, where 10\% of the collected data was reserved for the testing set, while the remaining 90\% was used for training. The data statistics are presented in Table \ref{tab:datastatistics}, providing an overview of the dataset composition. Additionally, Fig. \ref{fig:labelsample} showcases examples of labeled storefront objects within an image, providing a visual representation of the annotated data.

\begin{figure}[ht]
\begin{center}
   \includegraphics[width=0.8\linewidth]{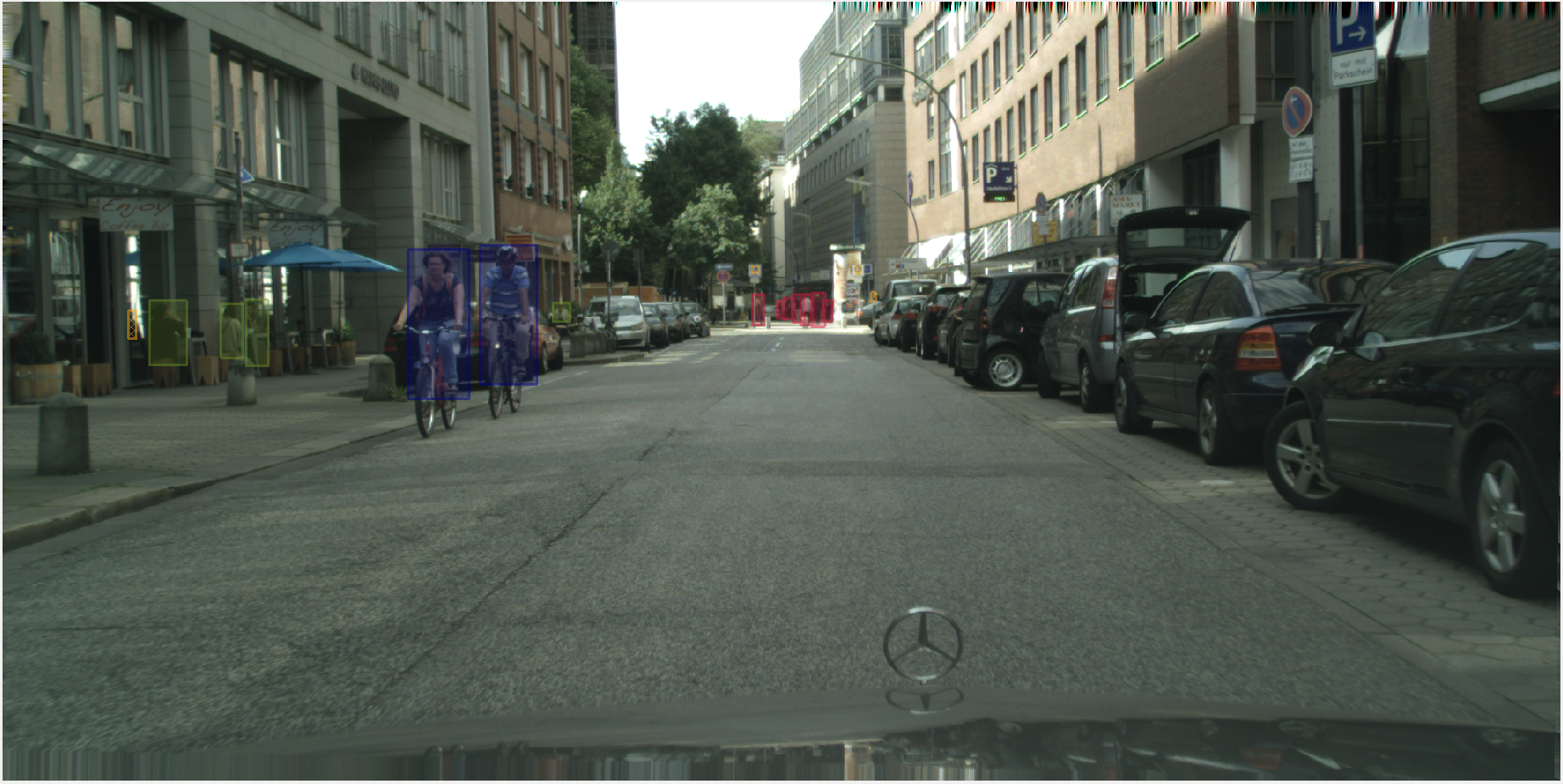}
\end{center}
   \caption{The label example from CityPersons Dataset \citep{zhang2017citypersons}. Red: Pedestrian. Blue: Rider. Yellow: Sitting person.}
\label{fig:citypersonexample}
\end{figure}

\textbf{CityPersons and CityPersons+ Dataset.} The CityPersons dataset is derived from the Cityscapes dataset \citep{Cordts2016Cityscapes}, focusing specifically on person annotations. It contains annotations for four categories: \textit{pedestrian, rider, sitting person}, and \textit{person (other)}. Table \ref{tab:citypersonstatistics} provides an overview of the dataset, including information on the number of images and annotations for each category. Figure \ref{fig:citypersonexample} showcases an example of labeled pedestrians from the dataset, providing a visual representation of the annotated data.

\begin{figure}[ht]
\begin{center}
   \includegraphics[width=0.8\linewidth]{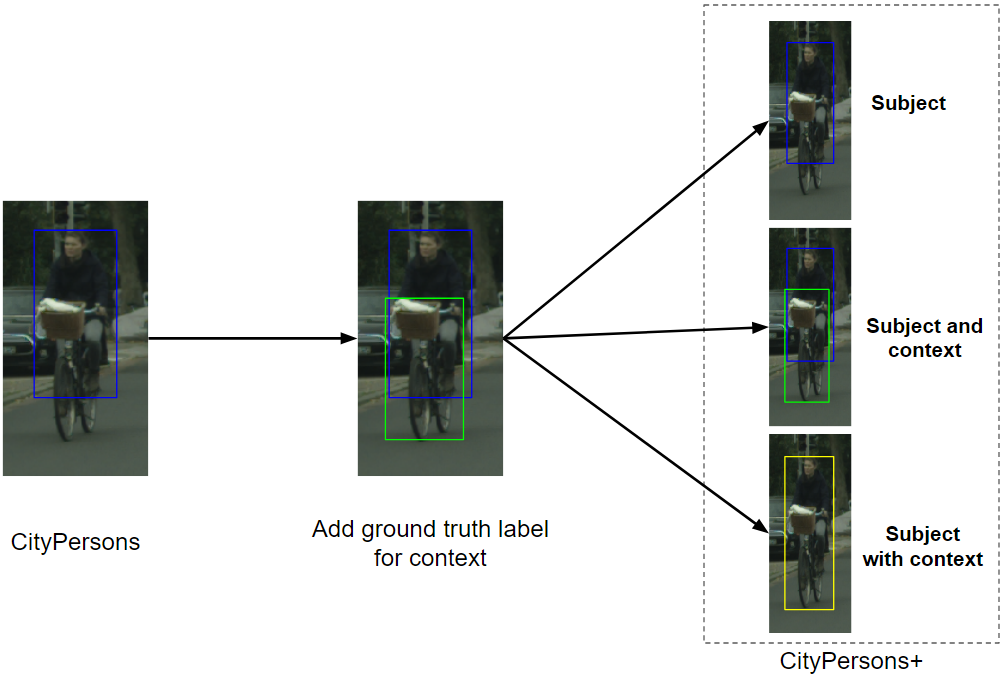}
\end{center}
   \caption{The demonstration of riders in CityPersons+ dataset. We extend existing categories in CityPersons dataset, with context information, by adding the ground truth label for context things and combined with the existing subject class label. }
\label{fig:citypersonplus}
\end{figure}

\begin{table}[ht]
  \centering
  \caption{Statistics of CityPersons and CityPersons+ Datasets.}
  \label{tab:citypersonstatistics}
  \resizebox{1\linewidth}{!}{\begin{tabular}{|c|c|c|c|}
      \hline
        Dataset & \# of Category & \# of Training & \# of Validation \\     \hline
        CityPersons \citep{zhang2017citypersons} & 4 & 2975 & 500 \\     \hline
        CityPersons+ & 6 & 2975 & 500 \\    \hline
  \end{tabular}
  }
\end{table}

To incorporate various context information and leverage the general topological relations between different categories, we introduce the CityPersons+ dataset. This dataset expands upon the CityPersons dataset by incorporating additional object labels from the Cityscapes dataset, including more specific subcategories. Specifically, we categorize pedestrians and riders into four subcategories: \textit{pedestrian on road, pedestrian on sidewalk, rider with motorcycle}, and \textit{rider with bicycle}. Therefore CityPersons+ contains annotations for six categories. The purpose of adding subcategories is to better utilizing context information. Fig \ref{fig:citypersonplus} shows how we include more context information without changing existing labels. We also relate the six categories in CityPersons+ dataset to context information that are beyond these six categories. First, we add the bounding box ground truth labels for \textit{context things}, including motorcycles, bicycles and vehicles, which are related to the existing subject class labels of rider with motorcycle, rider with bicycle, and pedestrian occluded by vehicle, respectively. Second, we include the the semantic segmentation labels of \textit{context stuff}, such as roads and sidewalks, which could provide precise spatial reasoning between different objects, namely, pedestrian on road, and pedestrian on sidewalk, in addition to pedestrian occluded by pedestrian. We also include word embeddings for both context things (motorcycles, bicycles and vehicles) and context stuff (roads and sidewalks) for Semantic Context Fusion (SCF) component. We use the pretrained model weights for Faster R-CNN and DETR to \textit{detect} the context things, and Segformer \citep{xie2021segformer} to \textit{segment} the semantic masks for context stuff, to facilitate general topological reasoning within the Spatial Contextual Reasoning (SCR) component (see Table \ref{tab:parametersettingssai}). Table \ref{tab:citypersonstatistics} provides an overview of the statistics for the CityPersons+ dataset, comparing with CityPersons dataset: we double the class categories for pedestrian and riders (from 2 to 4), add 5 context objects (not shown in the Table),  without changing the existing classes (2). For the 4 basic classes in CityPersons and 6 basic classes in CityPersons+, as shown in Table \ref{tab:citypersonstatistics}, the pretrained model weights for Faster R-CNN and DETR are finetuned using the two datasets, respectively, and the proposed GMC models will be evaluated.

\textcolor{black}{\textbf{MSCOCO-2017.} MSCOCO is a standard benchmark in object detection and instance segmentation. It includes 80 object categories with 118k images for training and 5k for evaluation. The dataset is known for its diversity, containing a wide range of objects and scenes. It features a maximum of 93 object instances per image, with an average of 7 objects.}

\subsection{Experimental Settings}
\label{experimentsettings}

\textbf{Faster R-CNN.} \textcolor{black}{In our implementation, we utilize ResNet-50 \citep{he2016deep} as the backbone feature extractor along with the Feature Pyramid Network (FPN) \citep{lin2017feature}, which are both pretrained on the COCO dataset.} For the semantic context fusion, we employ a 2-layer graph convolutional network (GCN) with LeakyReLU \citep{maas2013rectifier} as the activation function. The GCN takes 300-dimensional word embeddings from GloVe \citep{pennington2014glove} as the input label feature vector. During training, we employ Stochastic Gradient Descent (SGD) as the optimizer, with a momentum of 0.95 and a weight decay of 1e-4. The initial learning rate is set to 0.005 and is reduced by a factor of 0.25 every 8 epochs. We train the model for a total of 40 epochs for storefront accessibility detection, 60 epochs for pedestrian detection, and 50 epochs for COCO object detection.

\begin{table*}[ht]
    \color{black}\caption{Default user parameter settings for Spatial Context Reasoning in our experiments on the three datasets: SAI \citep{wang2022multiclu}, 
    CityPersons+, and COCO. O\_T: Overlap\_threshold.}
    \begin{center}
    \resizebox{0.85\linewidth}{!}{
    \color{black}\begin{tabular}{|c|c|c|c|c|c|c|c|}
    \hline
    Task & [Subject, Object] & Occlusion & Predicate & Topology & O\_T & Search\_area\_height & Search\_area\_width \\
    \hline
    \multirow{2}{*}{SAI } & [door, knob] & - & - & within & - & - & - \\
    & [door, stair] & - & under & overlap & 0.2 & 0.2$height_{door}$ + $height_{stair}$ & $width_{door}$ + $width_{stair}$ \\
    \hline
    \multirow{6}{*}{CityPersons+} & [rider, bicycle] & Reasonable & under & overlap & 0.48 & 0.5$height_{rider}$ & $width_{bicycle}$ \\
    
    & [rider, motorcycle] & Reasonable & under & overlap & 0.5 & 0.5$height_{rider}$ & $width_{motocycle}$ \\
    
     & [pedestrian, vehicle] & Heavy & under & overlap & 0.68 & - & - \\
    
     & [pedestrian, pedestrian] & Heavy & - & overlap & 0.76 & - & - \\
    
     & [pedestrian, road] & Reasonable & under & overlap & 0.2 & - & - \\
    
     & [pedestrian, sidewalk] & Reasonable & under & overlap & 0.13 & - & - \\
    \hline
    \multirow{9}{*}{COCO} & [person, person] & - & - & overlap & 0.73 & - & - \\
    & [person, surfboard] & - & under & overlap & 0.17 & 0.2$height_{person}$ & $width_{surfboard}$  \\
    & [person, tie] & - & - & within & - & - & - \\
    & [person, skateboard] & - & under & overlap & 0.1 & 0.2$height_{person}$ & $width_{skateboard}$ \\
    & [person, snowboard] & - & under & overlap & 0.16 & 0.2$height_{person}$ & $width_{snowboard}$ \\
    & [zebra, zebra] & - & - & overlap & 0.83 & - & -  \\
    & [baseball glove, person] & - & - & within & - & - & - \\
    & [potted plant, vase] & - & under & overlap & 0.45 & - & - \\
    & [frisbee, dog] & - & - & overlap & 0.85 & - & - \\
    \hline
    \end{tabular}
    }
    \end{center}
    \label{tab:parametersettingssai}
\end{table*}

\textbf{DETR.} Following the methodology described in \citep{carion2020end}, we utilize ResNet-50 as the feature extractor and a transformer encoder-decoder for our visual detector. The learning rate for both ResNet-50 and the transformer encoder-decoder is set to 0.005, and a weight decay of 1e-4 is applied. To train the model effectively, we set the maximum number of training epochs to 120 for storefront accessibility detection and 200 for pedestrian detection. During the training process, we log the results every 5 epochs, allowing for detailed monitoring of the model's performance and progress. These settings ensure a comprehensive and robust training process for achieving accurate detection results. 

\textcolor{black}{To ensure a fair comparison, we fine-tuned the pretrained parameters on COCO of the two baseline models on both SAI and CityPersons+ datasets. The configurations of the SCR component for the three tasks are shown in Table \ref{tab:parametersettingssai}.} 

\subsection{Experimental Results}
\label{experimentresults}
In this section, we present the comparison results for object detection on the SAI dataset (Section \ref{subOD}) and pedestrian detection on the CityPersons dataset (Section \ref{subPD}). We conduct comparisons with baseline detectors, including Faster R-CNN and DETR, as well as our previous context learning approaches \citep{wang2022multiclu, visapp23wang}, considering various combinations of our context learning and utilization components. The evaluation focuses on performance metrics such as precision, recall, and mean average precision (mAP), providing insights into the effectiveness of our proposed framework in enhancing object and pedestrian detection tasks.

To ensure a fair comparison between our proposed framework and the previously designed MultiCLU particularly for storefront accessibility detection\citep{wang2022multiclu}, we initially adopt the same settings as described in \citep{wang2022multiclu}. Specifically, we utilize the Small Object Dataset (SOD) standard to represent the local context for small objects in the SAI dataset. For this, we set the enlarge percentage to 15 percent, denoted as $\beta=0.15$. Similarly, we employ the same small object standard for the CityPersons dataset, with the enlarge percentage set to 10 percent, denoted as $\beta=0.10$. By using these consistent settings, we aim to facilitate a direct performance comparison between our proposed framework and MultiCLU.

\subsubsection{Storefront Object Detection}
\label{subOD}

\begin{table*}[ht]
    \caption{Comparison results on SAI dataset\citep{wang2022multiclu} with baseline detectors and previous context learning approaches. IT: Inference Time (s).}
    \begin{center}
        \resizebox{0.85\linewidth}{!}{
        \begin{tabular}{|c|c|c|c|c|c|c|c|c|c|}
        \hline
        \multirow{2}{*}{\textbf{Model}} & \multirow{2}{*}{\textcolor{black}{\textbf{IT}}} & \multicolumn{3}{|c|}{\textbf{Precision $\uparrow$}}  & \multicolumn{3}{|c|}{\textbf{Recall $\uparrow$}} & \multirow{2}{*}{\textbf{mAP $\uparrow$}} & \multirow{2}{*}{\textbf{Recall} $\uparrow$} \\\cline{3-8}
         & & Door & Knob & Stair & Door & Knob & Stair & &      \\ \hline
        Faster R-CNN \citep{ren2015faster} & 0.029 & 75.6 & 17.7 & 66.0 & 87.5 & 47.6 & 73.1 & 53.1 & 69.4 \\ \hline
        MultiCLU \citep{wang2022multiclu}  & 0.036 & 78.0 & 51.2 & 70.0  & 92.3 & 80.4 & 83.0 & 66.4 & 85.2        \\ \hline
        +LCR                 & 0.029 & 78.1 & 41.3 & 66.8            & 88.9 & 77.7 & 74.5  & 62.1 & 80.4     \\ \hline
        +SCF                 & 0.036 & 78.0 & 19.0 & 68.5            & 90.1 & 53.0 & 79.4 & 55.2 & 74.2        \\ \hline
        +SCR                 & 0.029 & 77.8 & 18.6 & 67.2            & 88.8 & 52.4 & 74.5 & 54.5 & 71.9       \\ \hline
        +LCR+SCF             & 0.036 & 78.4 & 50.0 & 69.2            & 90.8 & 75.0 & 79.4  & 65.9 & 81.7       \\ \hline
        +SCF+SCR             & 0.036 & 78.2 & 21.2 & 69.6            & 90.3 & 55.8 & 80.8 & 56.3 & 75.6        \\ \hline
        +LCR+SCR             & 0.029 & 79.2 & 41.2 & 67.8   & 89.2 & 77.8 & 74.5  & 62.7 & 80.5       \\ \hline
        GMC-C \citep{visapp23wang}  \& (this paper)     & 0.036 & 78.2 & 52.3 & 69.6   & 92.0 & 79.9 & 82.3  & 66.7 & 84.7      \\ \hline\hline
        DETR \citep{carion2020end} & 0.040 & 75.9 & 23.8 & 69.2 & 91.8 & 58.4 & 77.8 &  56.3 & 76.0\\ \hline
        +LCR                 & 0.040 & 77.0 & 45.6 & 68.5            & 90.5 & 75.4 & 79.4  & 63.7 & 81.7     \\ \hline
        +SCF                 & 0.045 & 77.8 & 27.6 & 70.0            & 91.4 & 61.5 & 81.2 & 58.5 & 78.0        \\ \hline
        +SCR                 & 0.040 & 77.4 & 25.2 & 69.6            & 90.8 & 60.8 & 79.0 & 57.4 & 76.9       \\ \hline
        +LCR+SCF             & 0.045 & 80.2 & 55.1 & 71.2            & 92.7 & 81.2 & 82.3  & 68.8 & 85.4       \\ \hline
        +SCF+SCR             & 0.045 & 78.2 & 29.8 & 69.2            & 91.4 & 62.3 & 81.5 & 59.1 & 78.4        \\ \hline
        +LCR+SCR             & 0.040 & 78.8 & 50.8 & 69.2   & 92.0 & 77.8 & 80.4  & 66.3 & 83.4       \\ \hline
        GMC-T (this paper)  & 0.045 & 80.6 & 55.8 & 71.2            &  92.7 &  82.0 &  82.6 & \textbf{69.2} &  \textbf{85.8}         \\ \hline
        \end{tabular}
        }
    \end{center}
    \label{tab:componentresultsai}
\end{table*}

In order to assess the effectiveness of our proposed general framework, we conducted a thorough comparison with two baseline detectors - Faster R-CNN \citep{ren2015faster} and DETR\citep{carion2020end},  and two of our previous context learning approaches \citep{wang2022multiclu, visapp23wang}, using the SAI dataset. Here we use MultiCLU to represent the specially designed multi-stage context framework with the CNN-based model Faster R-CNN, as reported in \citep{wang2022multiclu},  GMC-C to represent the GMC framework with the CNN-based model in this paper and also as reported in \citep{wang2022multiclu}, and GMC-T to represent the GMC framework on the DETR-based model. To gauge the effectiveness of our approach on small objects within the SAI dataset, we adopted the evaluation methodology outlined in \citep{wang2022multiclu}. Here, for the scenarios where the local context representation is employed, we leveraged both the original and expanded labels for small objects adhering to the defined criteria. In cases where both labels were detected for the same small object, we considered just one to eliminate any possibility of duplicate detections. The evaluation primarily focused on two key performance metrics: mean average precision (mAP) and recall. These metrics were measured at a standard Intersection over Union (IoU) threshold of 0.5, which is commonly used in object detection tasks.

\textbf{Performance comparison on Faster R-CNN\citep{ren2015faster}.} Our comparative analysis revealed significant performance improvements when applying our framework to the CNN-based models (represented in rows 1 to 3 of Table \ref{tab:componentresultsai}). Note for the SAI dataset, the GMC-C results have been reported in \citep{visapp23wang}, and the configuration is the same in this paper. Specifically, our GMC-C model outperformed Faster R-CNN, achieving substantial increases in both mAP (+13.6\%) and recall (+15.3\%). This highlights the effectiveness of our general context framework in enhancing object detection performance, surpassing the baseline detector. Furthermore, our GMC-C model exhibited a slightly higher mAP (+0.3\%) compared to the special MultiCLU model, which employed specialized context mechanisms. However, there was a slight decrease in recall (-0.5\%). 

The comprehensive comparison outcomes demonstrate the compelling performance of our framework when integrated into CNN-based models. By incorporating various context learning and utilization components, our framework successfully enhances both mAP and recall, surpassing the performance of baseline detectors and previous context learning approaches. This reaffirms the potential and value of our general context framework in advancing the field of computer vision and object detection tasks.

\textbf{Performance comparison on DETR \citep{carion2020end}.} To evaluate the flexibility and general applicability of our proposed framework, we extended its integration to the detection transformer architecture, represented by the DETR model \citep{carion2020end}. By incorporating the context learning components into the detection transformer, we conducted a comprehensive analysis of its impact on the detection performance. The evaluation results (rows 4 to 5 in Table \ref{tab:componentresultsai}) demonstrated significant improvements of our GMC-T model in both mean average precision (mAP) and recall compared to the baseline transformer model (DETR). Specifically, we observed a noteworthy increase of 12.9\% in mAP and 9.8\% in recall, highlighting the effectiveness of our context learning components in enhancing detection performance within the transformer framework. These findings further emphasize the adaptability and efficacy of our proposed framework, as it consistently improves detection performance across different model architectures. Note here that the transformer-based model already has context information learnt within the model, this is probably why the improvement (from DETR to GMC-T) is not as high as that on the CNN-based models (from Faster R-CNN to GMC-C). Nevertheless, the GMC-T model, which incorporates our context learning components into the detection transformer, emerged as the top-performing model among the evaluated configurations. This outcome underscores the versatility and effectiveness of our framework in enhancing detection capabilities across diverse model architectures, showcasing its potential for various object detection tasks.

Our proposed framework demonstrates superior performance on the SAI dataset, exhibiting significant improvements over the baseline detectors and delivering competitive results compared to our previous specially-designed context learning model MultiCLU \citep{wang2022multiclu}. These findings support the efficacy of our general context framework in improving object detection accuracy and recall rates, meanwhile adapting to different visual detector architectures. By efficiently leveraging contextual information, our framework enhances object detection accuracy and recall rates, demonstrating its flexibility and effectiveness in various detection scenarios.

\textbf{Performance comparison with different context components.} 
We embarked on a comprehensive performance comparison across various combinations of our three contextual components.The outcomes, presented in Table \ref{tab:componentresultsai}, illuminate compelling insights. 

First we analyze the performance improvements when using various combinations of contextual components on Faster RCNN. When each contextual component was applied in isolation, notable enhancements in recall (from 2.8\% to 11\%) and mAP (from 1.4\% to 9\%) over the baseline were discernible. Furthermore, it's intriguing to observe that when deploying individual contextual components, the impact of local contextual labeling was more pronounced than that of the other two components.

Upon considering combinations of two contextual components, a noteworthy trend emerged, with each combination outperforming the baseline detector. The improvements ranged from +3.2\% to 12.8\% for mAP and from 6.2\% to 12.3\% for recall. Strikingly, when the combinations encompassed the Local Context Representation (LCR) component, they exhibited substantial superiority over other combinations, showcasing considerable gains in both mAP (+6.4\% to 9.6\%) and recall (+4.9\% to 6.1\%). This outcome underscores the value of incorporating contextual information around small objects, notably accentuating the detection efficacy of vital elements like doorknobs. Moreover, in relation to the single LCR component, both Semantic Context Fusion (SCF) and Spatial Context Reasoning (SCR) exhibited positive impacts. These components further improved results over a single LCR component, influencing both mAP and recall positively. Intriguingly, when contrasting the application of both SCF and SCR against their individual application, the combined utilization marginally enhanced both mAP and recall compared to using them in isolation. 

The apex of our proposed framework's performance emerged with the integration of all three components (GMC-C), attaining a notable 13.6\% improvement in mAP and an impressive 15.3\% enhancement in recall over the baseline model Faster R-CNN. An interesting observation lies in the fact that our general framework enhances mAP across all categories in contrast to MultiCLU \citep{wang2022multiclu}, albeit with only minimal reductions in recall. This suggests that the specifically designed MultiCLU might introduce more false positives than accurate predictions, positioning our framework to offer heightened precision at the cost of slightly reduced recall.

One notable distinction between the two base models lies in the impact of the Local Context Representation (LCR) component. Specifically, the improvements achieved by using LCR with DETR are not as substantial as those observed with Faster R-CNN. \textcolor{black}{When solely applying the LCR component to Faster R-CNN, there is a remarkable enhancement in Precision and Recall for the "knob" category, with improvements of 23.6\% and 30.1\%, respectively. In contrast, when the LCR component is applied to DETR alone, the precision and recall see improvements of 21.8\% and 17.0\%, respectively, which are comparatively less effective than with Faster R-CNN. Moreover, the mAP and recall for Faster R-CNN see enhancements of 9.0\% and 11.0\%, whereas DETR experiences improvements of 7.4\% and 5.7\%, respectively, when the LCR component is added. This discrepancy could be attributed to the inherent self-attention mechanism of the transformer architecture, which inherently incorporates context information of local context especially for small objects, a feature that Faster R-CNN lacks. Nevertheless, the performance improvements achieved through various combinations of contextual components on DETR exhibit similar trends, indicating the consistent and robust functionality of the GMC framework across different backbone models.}


\subsubsection{Pedestrian Detection}
\label{subPD}

We conducted further evaluation of our general context learning and reasoning framework on pedestrian detection task using CityPersons dataset, comparing it with the baseline detectors, Faster R-CNN \citep{ren2015faster} and DETR \citep{carion2020end}, without any code modifications. Here again, we use GMC-C to represent the general framework of context learning with the CNN-based model, and GMC-T to represent the general framework on DETR-based model, on the original CityPersons dataset (without considering the subcategories or additional context for spatial context reasoning). In summary, in the labeling stage, we employ the small object standard for the CityPersons dataset to enhance the labeling of small objects with local context labeling. We further leverage the fine-grained category rider in CityPersons dataset to enable the semantic context fusion in the training stage, and the spatial context reasoning in the postprocessing stage. Note that the GMC-C model in this paper is the same as that in \citep{visapp23wang}.

Further, we use GMC-C+ and GMC-T+ to represent the general framework with more spatial context reasoning, using the CityPersons+ dataset with subcategories of pedestrians and riders, as well as information of vehicle, road and sidewalk. We compared the evaluation results on the \textit {reasonable} and \textit{heavy} subsets of the data using the standard evaluation metric in pedestrian detection, $MR^{-2}$ (where lower values indicate better performance). Here, the subsets were defined based on the height ($h$) and visible ratio ($v$) of pedestrians: Reasonable subset: $h \in [50, \infty]$, $v \in [0.65, 1]$; Heavy subset: $h \in [50, \infty]$, $v \in [0, 0.65]$.

\begin{table}[ht]
    \caption{Comparison results on Citypersons dataset\citep{zhang2017citypersons} with baseline detectors and previous context learning approaches. IT: Inference Time (s).}
    \begin{center}
        \resizebox{0.95\linewidth}{!}{
        \begin{tabular}{|c|c|c|c|}
        \hline
        \textbf{Model} & \textcolor{black}{\textbf{IT}} & \textbf{Reasonable $\downarrow$} & \textbf{Heavy $\downarrow$} \\ \hline
        Faster R-CNN\citep{ren2015faster} & 0.062 & 13.4 & 36.9 \\ \hline
        +LCR                 & 0.062 & 12.3 & 35.6     \\ \hline
        +SCF                 & 0.068 & 13.3 & 37.1         \\ \hline
        +SCR                 & 0.063 & 13.0 & 36.5        \\ \hline
        +LCR+SCF             & 0.068 & 12.2 & 35.2        \\ \hline
        +SCF+SCR             & 0.069 & 13.2 & 36.5         \\ \hline
        +LCR+SCR             & 0.063 & 12.0 & 36.0       \\ \hline
        GMC-C \citep{visapp23wang}\& (this paper)    & 0.069 & 12.0 & \textbf{35.2}  \\ \hline \hline
        DETR \citep{carion2020end} & 0.059 & 11.8 & 40.8 \\ \hline
        GMC-T (this paper)  & 0.063 & \textbf{10.5} & 39.5   \\ \hline
        \end{tabular}
        }
    \end{center}
    \label{tab:componentresultcp}
\end{table}

\textbf{Overall comparison with baseline detectors.} The comparison results presented in Table \ref{tab:componentresultcp} provide insights into the performance of the GMC framework on different architectures on both the reasonable and heavy subsets. It is observed that DETR and transformer-based GMC model (GMC-T) generally exhibits superior performance on the reasonable subset (+1.6\% and +2.9\%, respectively, compared to the Faster-RCNN base model), indicating its effectiveness in capturing contextual information and enhancing detection accuracy. However, DETR and GMC-T demonstrates lower performance on the heavy subset (-2.6\% and -3.9\% respectively, compared to the Faster-RCNN base model), which could be attributed to the absence of design elements such as the feature pyramid network (FPN) \citep{lin2017feature} employed in the Faster R-CNN framework. In contrast, the CNN-based model GMC-C may not achieve the same level of performance on the reasonable subset as transformer-based model GMC-T, but it often demonstrates 
better performance on the heavy subset (+1.7\% compared to the Faster-RCNN base model). This suggests that the CNN-based model are able to effectively handle challenging scenarios with heavily occluded pedestrians, where precise localization and robust feature extraction are crucial. This evidence supports our rationale of the general context framework in working with various backbone models depending on the task requirements.

\textbf{Performance comparison with different context components on Faster-RCNN.} 
Upon applying the Local Context Representation (LCR) component alone on Faster R-CNN, there was a noticeable enhancement of 1.1\% on the reasonable subset and 1.3\% on the heavy subset (as illustrated in Table \ref{tab:componentresultcp}). To further amplify our framework's capabilities, we introduced a fine-grained category (rider) into the CityPersons dataset during training to facilitate the Semantic Context Fusion (SCF) and Spatial Context Reasoning (SCR) components. As observed in the results analogous to those from the SAI dataset, configurations with the LCR component consistently yielded superior performance compared to other settings. However, it's worth noting that both SCF and SCR modules had a minor impact on pedestrian detection, possibly attributed to the relatively weak correlation between pedestrians and other urban objects. In summation, our comprehensive framework, encompassing all three components, achieved the most impressive performance across both the reasonable subset (1.4\% lower) and the heavy subset (1.7\% lower), outperforming the baseline detector and alternative combinations. 


\textbf{Comparison with DETR.} Upon comparing our newly introduced GMC-T model with the baseline Detection Transformer (DETR) model, our GMC-T model consistently demonstrated superior performance across both the "reasonable" and "heavy" subsets. This was marked by a substantial enhancement in detection performance, exhibiting an impressive 1.3\% improvement on both subsets. These results provide compelling evidence for the effectiveness of our context learning and reasoning components in bolstering the detection capabilities of diverse architectural frameworks. Moreover, our framework's adaptability is evident as it showcases its prowess not only in CNN-based models but also in transformer-based models. The ease with which our framework can be integrated and customized underscores its potential to cater to a range of visual detection tasks beyond just pedestrian detection.


Overall, the comparison results highlight the potential and versatility of our proposed context learning and reasoning components in improving object detection performance across different datasets and tasks. The framework offers a flexible and effective solution for incorporating context information and enhancing the detection capabilities of various deep learning models, contributing to advancements in the field of computer vision and object detection.

\begin{table}[ht]
    \caption{Comparison results on general spatial context reasoning (SCR) component with baseline detectors and previous designed component.}
    \begin{center}
        \resizebox{0.95\linewidth}{!}{
        \begin{tabular}{|c|c|c|}
        \hline
        \textbf{Model} & \textbf{Reasonable $\downarrow$} & \textbf{Heavy $\downarrow$} \\ \hline
        Faster R-CNN\citep{ren2015faster} & 13.4 & 36.9 \\ \hline
        Faster R-CNN + SCR & 12.8 & 36.1 \\ \hline
        GMC-C \citep{visapp23wang} \&(this paper)     & 12.0 & 35.2  \\ \hline
        GMC-C+  (this paper)   & 11.8 & \textbf{34.8}  \\ \hline
        DETR \citep{carion2020end} & 11.8 & 40.8 \\ \hline
        DETR + SCR & 11.2 & 39.8 \\ \hline
        GMC-T (this paper) &  10.5 & 39.5   \\ \hline
        GMC-T+ (this paper)  &  \textbf{10.2} & 38.6  \\ \hline
        \end{tabular}
        }
    \end{center}
    \label{tab:componentresultscrcomponent}
\end{table}

\textbf{The effectiveness of the general Spatial Context Reasoning (SCR).} We also conducted an extensive study to evaluate the effectiveness of the general spatial context reasoning (SCR) component within our framework. In order to achieve a more comprehensive and robust topological reasoning, we leveraged both bounding boxes for objects (such as bicycles, motorcycles, cars, pedestrians) and semantic masks for stuff (such as sidewalks and roads) in CityPersons+ dataset. This allowed us to capture and utilize the spatial relationships between various entities in the scene. To assess the impact of  the enhanced general SCR component, we evaluated its performance in two enhanced models - GMC-C+ and GMC-T+, as well as its use on the two baseline object detection models - Faster R-CNN and DETR. Table \ref{tab:componentresultscrcomponent} presents the comparative results of these models with and without the SCR component.

\textbf{(1). SCR performance on Faster R-CNN.} When we solely applied the SCR component to the Faster R-CNN model, we observed notable improvements in performance for both the reasonable and heavy subsets, achieving an increase of 0.6\% and 0.8\%, respectively. However, it is important to note that the Faster R-CNN model, without the inclusion of the local context and semantic context components, did not achieve the same level of performance as the GMC-C model. By replacing the initial spatial context reasoning component with our enhanced SCR component in the GMC-C model, leading to the GMC-C+ model, we observed a slight performance improvement of 0.2\% on the reasonable subset and 0.4\% on the heavy subset, over the GMC-C model. These results indicate that the integration of the enhanced SCR component can enhance the performance of the GMC-C model to some extent. However, when comparing these results with the performance of the enhanced SCR component alone (i.e., Faster R-CNN + SCR), it is evident that the GMC-C+ model with the combined local context, semantic context, and enhanced SCR component outperformed both subsets, achieving a significant improvement of 1.0\% on the reasonable subset and 1.3\% on the heavy subset. This demonstrates the synergistic effect of incorporating multiple context sources within the framework. our evaluation confirms that the integration of the enhanced general SCR component can effectively improve the performance of object detection models, particularly when combined with the local context and semantic context components. Overall, GMC-C+ achieves performance improvements of 1.6\% on the reasonable and 2.1\% on the heavy, compared to the Faster-RCNN base model.

\textbf{(2). SCR performance on DETR.} We also study whether our enhanced general SCR component can improve over the DETR model, which already incorporates a self-attention mechanism to leverage context information. Not surprisingly, even with the existing self-attention mechanism, the application of the enhanced SCR component to the DETR model led to performance improvements. Specifically, we observed an increase of 0.6\% on the reasonable subset and 1.0\% on the heavy subset, indicating that the SCR component can effectively enhance the context utilization capabilities of the DETR model. Furthermore, when we combined the general SCR component with the other two contextual components (local context and semantic context), our GMC-T+ model achieved additional performance improvements over the DETR model and the GMC-T model on both evaluation subsets. The results showed a significant improvement of 1.6\% on the reasonable subset and 2.2\% on the heavy subset, compared to the DETR base model, and a visible improvement of 0.3\% on the reasonable subset and 0.9\% on the heavy subset, compared to the GMC-T model. This highlights the complementary nature of the contextual components and their ability to further enhance the detection performance of the DETR model.

\begin{figure*}[ht]
\begin{center}
   \includegraphics[width=0.95\linewidth]{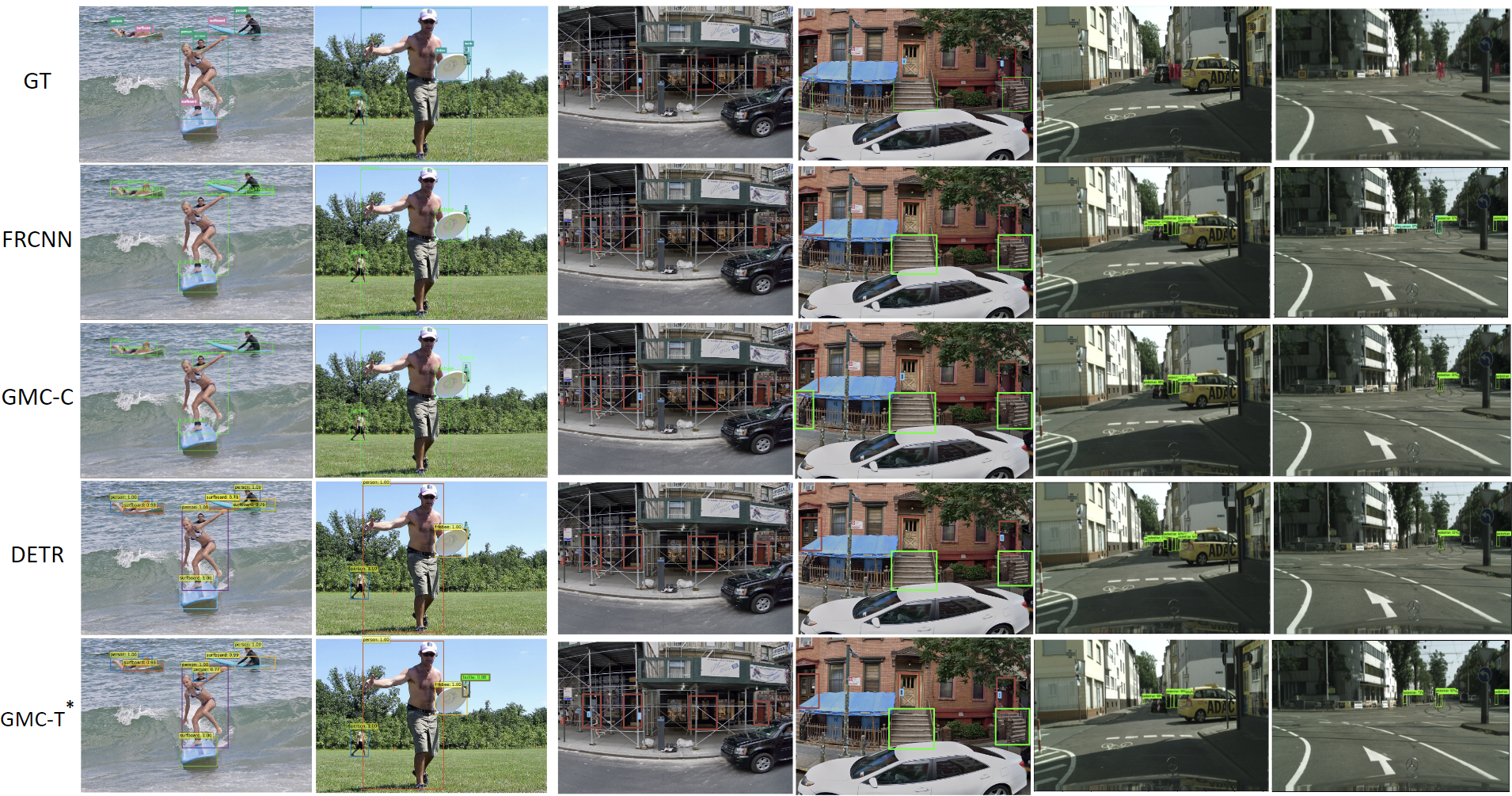}
\end{center}
   \color{black}\caption{Qualitative results on the three datasets: COCO  (columns 1 \& 2), SAI  (columns 3 \& 4)  and CityPersons+ (columns 5 \& 6). GMC-T*: We only evaluate the SCF and SCR components on COCO dataset, and the GMC-T was evaluated on the other two datasets.}
\label{fig:resultvis}
\end{figure*}

Our evaluation on pedestrian detection task confirms that the integration of the more general SCR component can effectively improve the performance of the detection models, particularly when combined with the local context and semantic context components. Our three contextual components, when integrated with the DETR model, demonstrated the best performance on the reasonable subset. On the other hand, the three contextual components combined with the CNN-based model Faster R-CNN exhibited better performance on the heavy subset. These findings indicate that the choice of model architectures, in combination with the specific context components, can have an impact on the overall detection performance, with different configurations achieving better results on different evaluation subsets. This also highlights the importance of leveraging multiple context sources and considering the spatial relationships between objects for achieving more accurate and robust detection. 

\subsection{\textcolor{black}{COCO Object Detection}}

\textcolor{black}{In order to check the scalability of our proposed general framework, we evaluate our framework on a large detection benchmark COCO dataset. We conducted comparison with two baseline detectors - Faster (\citep{ren2015faster}) and DETR (\citep{carion2020end}). We focus on two performance metrics: average precision (AP) and average precision for small objects ($AP_{S}$). The comparison results are shown in Table. \ref{tab:componentresultcoco}.}

\begin{table}[ht]
    \color{black}\caption{Comparison results on COCO dataset\citep{lin2014microsoft} with baseline detectors. IT: Inference Time (s).}
    \begin{center}
        \resizebox{0.95\linewidth}{!}{
        \color{black}\begin{tabular}{|c|c|c|c|}
        \hline
        \textbf{Model} & \textcolor{black}{\textbf{IT}} & \textbf{AP $\uparrow$} & \textbf{AP$_{S}$} $\uparrow$   \\ \hline
        Faster R-CNN \citep{ren2015faster} & 0.028 & 37.4 & 21.2 \\ \hline
        +LCR                 & 0.028 & 37.6 & 21.5     \\ \hline
        +SCF                 & 0.040 & 37.6 & 21.3        \\ \hline
        +SCR                 & 0.030 & 37.5 & 21.2       \\ \hline
        +LCR+SCF             & 0.030 & 37.9 & 21.6       \\ \hline
        +SCF+SCR             & 0.040 & 37.8 & 21.4        \\ \hline
        +LCR+SCR             & 0.028 & 37.7 & 21.6       \\ \hline
        GMC-C   & 0.040 & 38.1 & 21.7      \\ \hline\hline
        DETR \citep{carion2020end} & 0.036 & 42.0 & 21.0\\ \hline
        +SCF                 & 0.042 & 42.3 & 21.4        \\ \hline
        +SCR                 & 0.037 & 42.2 & 21.2      \\ \hline
        +SCF+SCR             & 0.042 & 42.7 & 21.5        \\ \hline
        \end{tabular}
        }
    \end{center}
    \label{tab:componentresultcoco}
\end{table}

\textcolor{black}{\textbf{Performance comparison on Faster R-CNN\citep{ren2015faster}.} 
Our comprehensive comparison results underscore the efficacy of our proposed GMC-C model, revealing significant improvements in key metrics. The average precision (AP) metric, a crucial indicator of overall detection performance, exhibited a notable enhancement of +0.7\% when employing our framework compared to the baseline Faster R-CNN. Moreover, our model demonstrated a noteworthy advancement in AP for small objects, registering an improvement of +0.5\%. This targeted improvement underscore the effectiveness of our proposed framework, particularly in addressing the detection challenges associated with smaller objects within the visual scene. The results substantiate the adaptability and enhanced performance of our GMC-C model, positioning it as a valuable asset in scenarios demanding precise and comprehensive object detection.}

\textcolor{black}{The application of the Local Context Representation (LCR) component in isolation on the Faster R-CNN model resulted in a modest improvement, with a 0.2\% increase in average precision (AP) and a 0.3\% enhancement in AP$_S$ (as detailed in Table \ref{tab:componentresultcoco}). Remarkably, when the LCR component was synergistically combined with the Semantic Context Fusion (SCF) component, this pairing exhibited the most substantial improvement compared to other combinations. The joint application yielded a 0.5\% boost in AP and a 0.4\% increase in AP$_S$. It is noteworthy that the individual application of the SCF and Spatial Context Reasoning (SCR) modules had a comparatively minor impact on the COCO dataset. In summary, our holistic framework, encompassing all three components, demonstrated the most remarkable performance improvement across both AP (+0.7\%) and AP$_S$ (+0.5\%), surpassing the baseline detector and alternative component combinations.}

\textcolor{black}{\textbf{Performance comparison on DETR \citep{carion2020end}.}
In our evaluation using DETR, the impact of our context components becomes apparent when applied individually. Since we have to fine-tune the large DETR model for LCR, we only tested performance improvements for the other two components (SCF and SCR) as the DETR can be frozen when training SCF and no re-training is needed for SCR. The Semantic Context Fusion (SCF) component, when introduced on its own, yields notable enhancements with a relative increase of +0.3\% on AP and +0.4\% on $AP_S$. This signifies that incorporating semantic relationships between objects contributes positively to the overall detection performance.}

\textcolor{black}{Conversely, the Spatial Context Reasoning (SCR) component, when applied independently, demonstrates a more modest impact, with only a +0.2\% improvement on both AP and $AP_S$. This result is suggestive of the challenges associated with defining meaningful relations between objects in the COCO dataset, where the provided relations are limited.}

\textcolor{black}{Interestingly, the synergy between SCF and SCR components becomes evident when they are combined. Their complementary nature enhances each other's contributions, resulting in a more substantial improvement. The joint application of SCF and SCR leads to a further increase in performance, with a +0.7\% improvement on AP and +0.5\% on $AP_S$. This collaborative effect underscores the value of integrating both semantic and spatial context reasoning for more effective object detection within the DETR framework.}

\subsection{\textcolor{black}{Performance discussions for different tasks/datasets}}

\textcolor{black}{With more in-depth examinations, we sought to delineate the specific object categories that exhibit significant influence from the Spatial Context Reasoning (SCR) component across the diverse datasets we scrutinized. As shown in Table. \ref{tab:impactedcattable}, within the SAI dataset, the SCR component dynamically integrates contextual relationships for all three categories—door, knob, and stair. Transitioning to the CityPersons+ dataset, the SCR component extends its reach across the entire spectrum of object categories. Notably, contextual elements like road and sidewalk draw upon insights from a state-of-the-art segmentation model, leading to a pronounced impact on 75\% of the dataset's categories. In the case of the COCO dataset, the SCR component centers its focus on the person category, given its preeminence as the most abundant class in the dataset. While other categories also experience influence, the overall impact encompasses approximately 13.75\% of all object categories within the COCO dataset.}  

\begin{table}[ht]
    \color{black}\caption{Impacted Categories for all datasets in SCR component.}
    \begin{center}
        \resizebox{0.85\linewidth}{!}{
        \color{black}\begin{tabular}{|c|c|c|}
        \hline
        \textbf{Datasets} & \textbf{Impacted Categories} & \textbf{Percentage} \\ \hline
        SAI & 3/3 & 100 \\ \hline
        Citypersons+ & 6/8 & 75 \\ \hline
        COCO     & 11/80 & 13.75  \\ \hline
        \end{tabular}
        }
    \end{center}
    \label{tab:impactedcattable}
\end{table}

\begin{table}[ht]
    \color{black}\caption{Component performance on most impacted categories on all dataset. D:DETR. F:Faster R-CNN.}
    \begin{center}
    \resizebox{1\linewidth}{!}{
    \color{black}\begin{tabular}{|c|c|c|c|c|c|}
    \hline
    Dataset & Category & Model & AP $\uparrow$ &  Reasonable$\downarrow$ & Heavy$\downarrow$  \\
    \hline
    SAI & knob & D+LCR & 23.8 $\rightarrow$ 45.6 
    & - & - \\ \hline
    CityPersons+ & pedestrian & F+LCR &  - & 13.4 $\rightarrow$ 12.3 & 36.9 $\rightarrow$ 35.6 \\ \hline
    COCO & person & D+SCR & 47.3 $\rightarrow$ 50.9 & - & - \\\hline
    
    \end{tabular}
    }
    \end{center}
    \label{tab:componentresultoncategories}
\end{table}

\textcolor{black}{We further conducted evaluations to assess how our components perform on the most impacted categories across all datasets, and the summarized results are presented in Table \ref{tab:componentresultoncategories}. In the SAI dataset, the substantial improvement of +21.8\% in AP for the "knob" category, achieved by applying the Local Context Representation (LCR) component with DETR, underscores the pivotal role of contextual information in detecting and delineating small objects. This result suggests that leveraging local context in tandem with transformer-based models significantly benefits the identification of intricate details in specific categories. Moving to the CityPersons+ dataset, where the "pedestrian" category exhibited the most notable enhancement of +1.1\% on the reasonable set and +1.3\% on the heavy set with the LCR component on Faster R-CNN, we observe the importance of local context in urban scenes. The improved detection performance for pedestrians, a crucial element in urban scenarios, emphasizes the significance of considering context for specific object classes. This insight becomes especially valuable in the domain of object detection, where capturing fine-grained details is essential.}

\textcolor{black}{In the COCO dataset, the "person" category's substantial improvement of 3.6\% in AP with the Spatial Context Reasoning (SCR) component applied to the DETR model suggests that accounting for spatial relationships is particularly beneficial in datasets characterized by a larger scale and diverse object categories. Spatial reasoning plays a crucial role in refining the predictions, especially in scenarios where objects interact in complex spatial configurations. Although Semantic Context Fusion (SCF) didn't exhibit standout improvements compared to the other two components, its role in contributing to enhanced performance, especially when combined with LCR and SCR components, underscores its potential in capturing contextual semantics. This holistic approach, leveraging different forms of context throughout the entire deep learning process, demonstrates promising results and sets the stage for further exploration in context-aware computer vision tasks.}

\textcolor{black}{Furthermore, we conducted a thorough comparison of the inference times (expressed in seconds) across our results(Tables \ref{tab:componentresultsai}, \ref{tab:componentresultcp} and \ref{tab:componentresultcoco}). The findings revealed that our framework incurs only a marginal increase in time complexity. Furthermore, the qualitative results visualized in Figure \ref{fig:resultvis} provide a compelling illustration of how the proposed method enhances performance across all three datasets (COCO, SAI, and CityPersons+), offering a comprehensive validation of its efficacy.}

\section{Conclusions and Discussion}\label{conclusion}

In summary, we have proposed a general framework of multistage context learning and utilization for visual detection tasks. Our proposed framework consists three context components to utilize local context, semantic context and spatial context information. The three context components have the flexibility and adaptability to utilize the framework across various visual detection tasks, with different visual detectors. The proposed framework are evaluated and verified on complex street scenes for a storefront object detection task and a pedestrian detection task. Compared to the state of the art methods, the evaluation demonstrates that our framework can efficiently leveraging contextual information at various stages such as data preprocessing, model training and post-processing. Our comparison results also show that the proposed contextual components can effectively improve the performance over different baseline models, with the support of different context information.

However, there is still space for improvements over the proposed framework. In this work, we only explore local, global and semantic context, mostly in the spatial domain. Other context types need more attention, and new architectures particularly designed for context learning and utilization as summarized in \citep{wang2023context} have not been considered. 

\textcolor{black}{Despite our attempt in conducting experiments on the extensive MSCOCO dataset to show promising results, defining general spatial relations of all object categories becomes a challenge, especially when dealing with a dataset that encompasses numerous categories. The task of establishing meaningful and universally applicable spatial relations becomes intricate due to the diversity of object categories present in the dataset. Addressing this challenge requires a thoughtful approach to derive spatial relations that can effectively generalize across a wide range of object types. Further exploration and research may be needed to develop a robust and adaptable method for defining spatial relations that accommodates the inherent diversity of categories within the dataset.} 

Furthermore, there are many works focus on the real world detection scenarios, where the standard evaluation metrics may not work well. A contextual evaluation based on the requirements of real-world applications is needed not only for object detection task, but may also benefit other computer vision tasks. 
 
\section*{Acknowledgements}
\label{sec:acknowledgements}
The work is supported by the National Science Foundation (NSF) through Awards \#2131186 (CISE-MSI),  \#1827505 (PFI), and \#1737533 (S\&CC). The work is also supported by the US Air Force Office of Scientific Research (AFOSR) via Award \#FA9550-21-1-0082, a College-wide Research Vision (CRV) Fund from the CCNY Provost's Office, and the ODNI Intelligence Community Center for Academic Excellence (IC CAE) at Rutgers University (\#HHM402-19-1-0003 and \#HHM402-18-1-0007).

\bibliographystyle{model2-names}
\bibliography{references.bib}



\end{document}